%% file: article.tex
\documentclass[10pt,journal,compsoc]{IEEEtran}
\usepackage[pdftex]{graphicx}
\usepackage{subcaption}
\usepackage{amssymb,amsmath}
\usepackage{amsthm} 
\usepackage{mathtools}
\usepackage[table]{xcolor}
\usepackage{multirow}
\usepackage{hyperref}
\usepackage{enumerate}
\usepackage{booktabs}
\usepackage[utf8]{inputenc}
\usepackage{array}
\usepackage[textsize=tiny, textwidth=1.3cm]{todonotes}
\usepackage[stable]{footmisc}
\usepackage{bbm}
\usepackage{float}
\usepackage{tikz} 
\usetikzlibrary{bayesnet} 
\usepackage[ruled,vlined,linesnumbered]{algorithm2e}

\definecolor{deep_blue}{rgb}{0,.2,.5}
\definecolor{dark_blue}{rgb}{0,.15,.5}

\hypersetup{pdftex,  
	breaklinks=true,  
	colorlinks=true,
	linkcolor=dark_blue,
	citecolor=deep_blue,
}

\newcommand\norm[1]{\lVert#1\rVert}

\newcolumntype{L}[1]
{>{\raggedright\let\newline\\\arraybackslash\hspace{0pt}}m{#1}}
\newcolumntype{C}[1]
{>{\centering\let\newline\\\arraybackslash\hspace{0pt}}m{#1}}
\newcolumntype{R}[1]
{>{\raggedleft\let\newline\\\arraybackslash\hspace{0pt}}m{#1}}

\newtheorem{proposition}{Proposition}[section]
\newtheorem{theorem}{Theorem}[section]

\SetKwInOut{Input}{Input}
\SetKwInOut{Output}{Output}

\ifCLASSOPTIONcompsoc
\usepackage[nocompress]{cite}
\else
\usepackage{cite}
\fi
\ifCLASSINFOpdf
\usepackage[pdftex]{graphicx}
\else
\fi

\usepackage{stfloats}
\usepackage{url}

\begin{document}
	%
	\title{Encoding high-cardinality \\string categorical variables}
	%
	%
	%
	%
	
	\author{Patricio~Cerda
		and~Gaël~Varoquaux
		\IEEEcompsocitemizethanks{
			\IEEEcompsocthanksitem Inria, Parietal team \protect\\
			E-mail: patricio.cerda@inria.fr
		}
	}

	\IEEEtitleabstractindextext{%
		\begin{abstract}
			Statistical models usually require vector
			representations of
			categorical variables, using for instance \emph{one-hot encoding}.
			This strategy breaks down
			when the number of categories grows, as it
			creates high-dimensional feature vectors. Additionally,
			for string entries, one-hot encoding does not
			capture morphological information in their representation.

			Here, we seek low-dimensional encoding of high-cardinality string
			categorical variables. Ideally, these should be:
			scalable to many categories;
			interpretable to end users; and
			facilitate statistical analysis.
			We introduce two encoding approaches for string
			categories: a \textit{Gamma-Poisson matrix factorization}
			on substring counts,
			and a \textit{min-hash encoder}, 
			for fast
			approximation of string similarities.
			We show that min-hash turns set inclusions
			into inequality relations that are easier to learn.
			Both approaches are scalable and streamable.
			Experiments on real and simulated data show that these
			methods improve
			supervised learning
			with high-cardinality categorical variables.
			We recommend the following:
			if scalability is central,
			the min-hash encoder is the best option as it does not
			require any data fit;
			if interpretability is important, the Gamma-Poisson
			factorization is the best alternative,
			as it can be interpreted as one-hot encoding on
			inferred categories with informative feature names.
			Both models enable autoML on string
			entries as they remove the need for
			feature engineering or data cleaning.
		\end{abstract}

		\begin{IEEEkeywords}
		Statistical learning, string categorical variables, autoML,
		interpretable machine learning, large-scale data, min-hash,
		Gamma-Poisson factorization.
		\end{IEEEkeywords}
	}

	\maketitle

	\IEEEdisplaynontitleabstractindextext

	%
	\IEEEpeerreviewmaketitle
	
	\IEEEraisesectionheading{\section{Introduction}\label{sec:introduction}}
	
	\IEEEPARstart{T}{abular} datasets often contain columns with string entries.
	However, fitting statistical models on such data generally requires
	a  numerical representation of all entries, which calls for building an
	\emph{encoding}, or vector representation of the entries.
	Considering string entries as nominal---unordered---categories
	gives well-framed statistical analysis. In such situations, categories
	are assumed to be mutually exclusive and unrelated, with a fixed known
	set of possible values.
	Yet, in many real-world datasets, string columns are not
	standardized in a small number of categories.
	This poses challenges for statistical analysis.
	First, the set of all possible categories may be huge and not known a
	priori, as the
	number of different strings in the column can indefinitely increase with the number of samples.
	Second, categories may be related: they often carry
	some morphological or semantic links.
	
	The classic approach to encode categorical variables for statistical
	analysis is \emph{one-hot encoding}. It
	creates vectors that agree with the general intuition of
	nominal categories: orthogonal and equidistant
	\cite{cohen2013applied}.
	However, for high-cardinality categories, one-hot encoding leads to
	feature vectors of high dimensionality.
	This is especially problematic in big data settings, which can lead to a
	very large number of categories, posing computational and
	statistical problems. 
	
	Data engineering practices typically tackle these issues with 
	data-cleaning techniques \cite{pyle1999data,rahm2000data}. In
	particular, deduplication
	tries to merge different variants of the same entity
	\cite{winkler2006overview,elmagarmid2007duplicate,christen2012data}.
	A related concept is that of \emph{normalization}, used in databases and
	text processing to put entries in canonical forms.
	However, data cleaning or normalization often requires human intervention,
	and
	are major costs in data analysis\footnote{
		Kaggle industry survey: \url{https://www.kaggle.com/surveys/2017}}.
	To avoid the cleaning step, \emph{Similarity encoding}
	\cite{cerda2018similarity} relaxes one-hot encoding by using
	\emph{string similarities} \cite{gomaa2013survey}.
	Hence, it addresses the problem of related categories and has been shown
	to improve statistical analysis upon one-hot encoding
	\cite{cerda2018similarity}.
	Yet, it does not tackle the problem of high cardinality, and the data
	analyst much resort to heuristics such as choosing a subset of the
	training categories
	\cite{cerda2018similarity}.
	
	Here, we seek encoding approaches for statistical analysis on
	string categorical entries that are suited to a very large number of
	categories without any human intervention: avoiding data cleaning,
	feature engineering, or neural architecture search. Our goals are: 
	\emph{i)} to provide feature vectors of limited dimensionality without any
	cleaning or feature engineering step, even for very large datasets;
	\emph{ii)} to improve statistical analysis tasks such as supervised
	learning; and \emph{iii)} to preserve the intuitions behind
	categories: entries can be arranged in natural groups that can be
	easily interpreted.
	We study two novel encoding methods that both address scalability and
	statistical performance: a \textit{min-hash encoder}, based on 
	locality-sensitive hashing (LSH) \cite{gionis1999similarity},
	and a low-rank model of co-occurrences in character n-grams:
	a \textit{Gamma-Poisson matrix factorization}, suited to counting statistics.
	Both models scale linearly with the number of samples and are suitable
	for statistical analysis in streaming settings.
	Moreover, we show that the Gamma-Poisson factorization model
	enables interpretability with a sparse encoding that expresses the
	entries of the data as linear
	combinations of a small number of latent categories,
	built from their substring information.
	This interpretability is very important: opaque and
	black-box machine learning models have limited adoption in
	real-world data-science applications. Often, practitioners resort to manual
	data cleaning to regain interpretability of the models. 
	Finally, we demonstrate on 17 real-life datasets that
	our encoding methods improve supervised learning on non curated data
	without the need for dataset-specific choices.
	As such, these encodings provide a scalable and automated replacement
	to data cleaning or feature engineering, and restore the benefits
	of a low-dimensional categorical encoding, as one-hot encoding.
	
	The paper is organized as follows.
	Section \ref{sec:previous_work} states the problem and the
	prior art on creating feature vectors from categorical variables.
	Section \ref{sec:scalable_encoders} details our two
	encoding approaches.
	In section \ref{sec:experiments}, we present our experimental study
	with an emphasis on interpretation and on statistical learning
	for 17 datasets with non-curated entries and 7 curated ones.
	Section \ref{sec:discussion} discusses these results, after which
	appendices provide information on the datasets and the experiments to
	facilitate the reproduction of our findings.

	\section{Problem setting and prior art}
	\label{sec:previous_work}
	
	The statistics literature often considers datasets that contain
	only categorical variables with a low cardinality, as
	datasets\footnote{See for example, the Adult dataset (%
		\url{https://archive.ics.uci.edu/ml/datasets/adult})} in the UCI repository
	\cite{dua2017uci}. 
	In such settings, the popular \emph{one-hot encoding} is a suitable solution
	for supervised learning \cite{cohen2013applied}: it models categories as
	mutually exclusive and, as categories are known a priory,
	new categories are not expected to appear in the test set.
	With enough data, supervised learning can then be used to link each
	category to a target variable. 
	
	\subsection{High-cardinality categorical variables}
	
\begin{figure}[t!]
		\hspace{1cm}
		\centering
		\includegraphics[trim={21cm 8.5cm 0cm .1cm},clip,width=.46\textwidth]
			{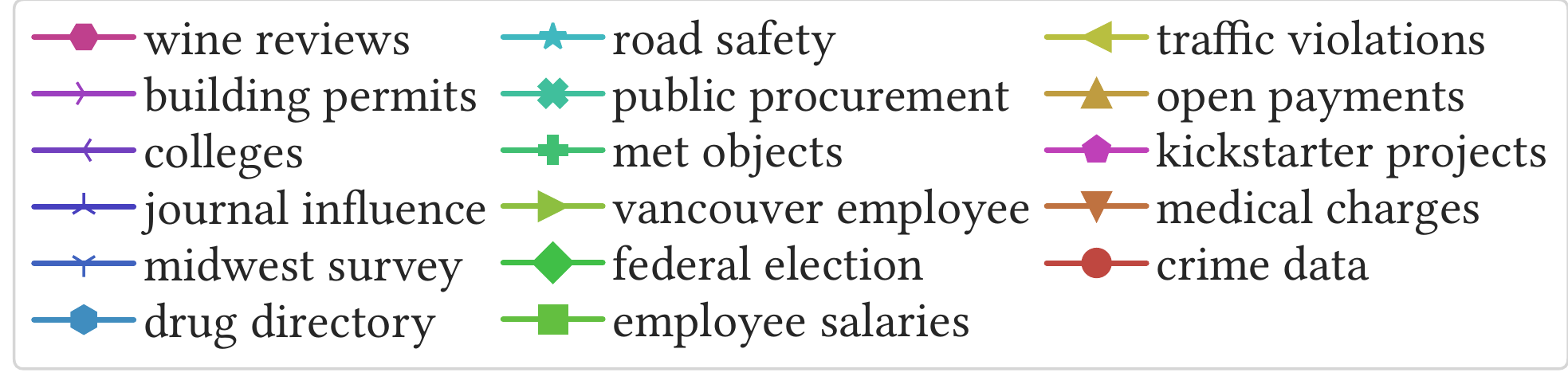}
			
		\includegraphics[trim={0 0 20cm 0},clip,height=.2\textheight]
		{datasets_unique_categories}
		\caption{\textbf{Number of categories
				versus number of samples}. In general,
			a higher number of samples implies a higher number of categories for
			the respective variable. In general, the cardinality of categories
			grows slower than words in a typical NLP problem
			(Wikipedia articles in this case).}
		\label{fig:cats_vs_nsamples}
	\end{figure}

	However, in many real-world problems, the number of different string
	entries in a column is very large, often growing with the number of
	observations (\autoref{fig:cats_vs_nsamples}). 
	Consider for instance the \emph{Drug Directory} dataset\footnote{
		Product listing data 
		for all unfinished, unapproved drugs.
		Source: U.S. Food and Drug Administration (FDA)}.
	One of the variables is a categorical column with
	\emph{non proprietary names} of drugs.
	As entries in this column have not been normalized, many
	different entries are likely related: 
	they share a common ingredient such as \texttt{alcohol} (see \autoref{subtab:drug_directory}).
	Another example is the \emph{Employee Salaries} dataset\footnote{
		Annual salary information for employees of the Montgomery County, MD, U.S.A.
		Source: \url{https://data.montgomerycountymd.gov/}}.
	Here, a relevant variable is the \textit{position title} of employees.
	As shown in \autoref{subtab:employee_salaries}, here there is also
	overlap in the different occupations.
	
	High-cardinality categorical variables may arise from 
	variability in their string representations,  such as 
	abbreviations, special characters, or typos\footnote{A
		taxonomy of different sources of \emph{dirty data}
		can be found on~\cite{kim2003taxonomy}, and a formal description of
		data quality problems is proposed by \cite{oliveira2005formal}.}.
	Such non-normalized data often contains very rare categories.
	Yet, these categories tend to have common morphological information. 
	Indeed, the number of unique entries grows less fast with the size of the
	data than the number of words in natural language
	(\autoref{fig:cats_vs_nsamples}).
	In both examples above, drug names
	and position titles of employees, there is an implicit taxonomy.
	Crafting feature-engineering or
	data-cleaning rules can recover a small number of relevant categories.
	However, it is time consuming and often needs domain expertise.

	\paragraph*{\bf Notation} We write sets of elements with capital curly fonts,
	as $\mathcal{X}$. Elements of a vector space (we consider row vectors)
	are written in bold $\mathbf{x}$ with the $i$-th entry denoted by
	$x_{i}$, and matrices are in capital and bold $\mathbf{X}$, with $x_{ij}^{}$
	the entry on the $i$-th row and $j$-th column.
	
	\begin{table}[t]
		\caption{Examples of high-cardinality categorical variables.%
			\label{tab:high-cardinality-categorical_variables}}
		\begin{subtable}[t!]{.52\linewidth}	
			\rowcolors{2}{black!5}{white}
			\begin{tabular}{r l}	
				\hline\noalign{\smallskip}
				\textbf{Count} & \textbf{Non Proprietary Name}       \\
				\noalign{\smallskip}\hline\noalign{\smallskip}
				1736 &  \textbf{alcohol}  \\
				1089 &  ethyl \textbf{alcohol}  \\
				556 &  isopropyl \textbf{alcohol}  \\
				16 &  polyvinyl \textbf{alcohol}  \\
				12 &  isopropyl \textbf{alcohol} swab  \\
				12 &  62\% ethyl \textbf{alcohol}  \\
				6 &  \textbf{alcohol} 68\%  \\
				6 &  \textbf{alcohol} denat  \\
				5 &  dehydrated \textbf{alcohol}  \\
				\hline\noalign{\smallskip}
			\end{tabular}
			\caption{Count for some of the categories containing the word
				\emph{alcohol} in the \emph{Drug Directory} dataset.
				The dataset contains more than 120k samples.}
			\label{subtab:drug_directory}
		\end{subtable}%
		\hfill
		\begin{subtable}[t!]{.44\linewidth}
			\rowcolors{2}{black!5}{white}
			\begin{tabular}{l}
				\hline\noalign{\smallskip}
				\textbf{Employee Position Title}\\
				\noalign{\smallskip}\hline\noalign{\smallskip}
				\textbf{Police} Aide	\\
				Master \textbf{Police} Officer	\\
				Mechanic \textbf{Technician} II\\
				\textbf{Police} Officer III	\\
				Senior Architect\\
				Senior Engineer \textbf{Technician}\\
				Social Worker III	\\
				Bus Operator \\
				\hline\noalign{\smallskip}
			\end{tabular}
			\caption{Some categories in the \emph{Employee Salaries} dataset.
				For 10\,000 employees, there are 
				almost 400 different occupations. Yet,
				they share relevant substrings.}
			\label{subtab:employee_salaries}
		\end{subtable}
	\end{table}

	Let $C$ be a categorical variable such that
	$\text{dom}(C) \subseteq \mathcal{S}$, the set of finite length strings.
	We call \textit{categories} the elements of $\text{dom}(C)$.
	Let $s_i {\in} \mathcal{S}, i{=}1...n$, be the category
	corresponding to the $i$-th sample of a dataset. For statistical learning,
	we want to find an encoding function
	$\text{enc:}\, \mathcal{S} \rightarrow \mathbb{R}^d$,
	such as $\text{enc}(s_i) = \mathbf{x}_i$. We call $\mathbf{x}_i$ the
	\textit{feature map} of $s_i$. \autoref{tab:notations_summary}
	contains a summary of the main variables used in the next sections.

	\subsection{One-hot encoding, limitations and extensions}
	
	\subsubsection{Shortcomings of one-hot encoding}
	
	From a statistical-analysis standpoint, the
	multiplication of entries with related information is challenging
	for two reasons. First, it dilutes the information: learning on rare
	categories is hard. Second, with one-hot encoding, representing these as
	separate categories creates high-dimension feature vectors.
	This high dimensionality entails large computational and memory costs;
	it increases the complexity of the associated learning problem,
	resulting in a poor statistical estimation \cite{bottou2008tradeoffs}.
	Dimensionality reduction of the one-hot encoded matrix can help with
	this issue, but at the risk of loosing information. 
	
	Encoding all unique entries with orthogonal vectors discards the
	overlap information visible in the string representations.
	Also, one-hot encoding cannot
	assign a feature vector to new categories that may appear in the
	testing set, even if its representation is close to one in the training set.
	Heuristics such as assigning the zero vector to new categories,
	create collisions if more than one new category appears.
	As a result, one-hot encoding is ill suited to online learning settings:
	if new categories arrive, the entire encoding of the dataset has
	to be recomputed and the dimensionality of the feature vector becomes
	unbounded.
	
	\begin{table}[t!]
		\caption{Summary of notations}
		\label{tab:notations_summary}
		\setlength\tabcolsep{5pt}
		\rowcolors{2}{black!5}{white}
		\begin{tabular}{ll}
			\toprule
			\textbf{Symbol} & \textbf{Definition} \\
			\midrule
			$\mathcal{S}$ & Set of all finite-length strings. \\
			$\mathcal{G}(s) \subseteq \mathcal{S}$ &
			Set of all consecutive n-grams in $s \in \mathcal{S}$. \\
			$\mathcal{V} = \bigcup_{i=1}^n \mathcal{G}(s_i)$ &
			Vocabulary of n-grams in the train set. \\
			$C$ & Categorical variable. \\
			$n$ & Number of samples. \\
			$d$ & Dimension of the categorical encoder. \\
			$m = |\mathcal{V}|$ & Cardinality of the vocabulary. \\	
			$\mathbf{F} \in \mathbb{R}^{n \times m}$ &
			Count matrix of n-grams. \\
			$\mathbf{X} \in  \mathbb{R}^{n \times d}$ &
			Feature matrix of $C$. \\
			$\text{sim}: \mathcal{S} \times \mathcal{S} \rightarrow [0, 1]$ &
			String similarity. \\
			$h_k: \mathcal{S} \rightarrow [0, 1]$ & Hash function with salt value equal to $k$.\\
			$Z_k: \mathcal{S} \rightarrow [0, 1]$ & Min-hash function with salt value equal to $k$.\\
			\hline
		\end{tabular}
	\end{table}
	
	\subsubsection{Similarity encoding for string categorical variables}
	
	For categorical variables represented by strings,
	\emph{similarity encoding} extends one-hot encoding by taking into account
	a measure of string similarity between pairs of categories
	\cite{cerda2018similarity}.
	
	Let $s_i {\in} \mathcal{S}, i{=}1...n$, the category
	corresponding to the $i$-th sample of a given training dataset.
	Given a string similarity
	$\text{sim}(s_i, s_j){:}\, \mathcal{S}{\times}\mathcal{S} {\rightarrow} [0, 1]$,
	similarity encoding
	builds a feature map $\mathbf{x}_i^{\text{sim}} {\in} \mathbb{R}^k$ as:
	\begin{equation}
	\mathbf{x}_i^{\text{sim}} \overset{\text{def}}{=} [ \text{sim}(s_i^{}, s^{(1)}),
	\text{sim}(s_i^{}, s^{(2)}), \dots,
	\text{sim}(s_i^{}, s^{(k)})] \in \mathbb{R}^k,
	\end{equation} 
	where $\lbrace s^{(l)}, l{=}1 \dots k \rbrace \subseteq \mathcal{S}$
	is the set of all unique categories in the train set---or a subset
	of prototype categories chosen heuristically\footnote{
		In this work, we use as dimensionality reduction technique
		the k-means strategy explained in \cite{cerda2018similarity}.}.
	With the previous definition, one-hot encoding corresponds to taking
	the discrete string similarity:
	\begin{equation}
	\text{sim}_{\text{one-hot}} (s_i^{}, s_j^{}) = \mathbbm{1}[s_i^{} = s_j^{}],
	\end{equation}
	where $\mathbbm{1}[\cdot]$ is the indicator function.
	
	Empirical work on databases with categorical columns containing
	non-normalized entries showed that similarity encoding with a continuous
	string similarity brings significant benefits upon one-hot encoding
	\cite{cerda2018similarity}. Indeed, it relates rare categories to
	similar, more frequent ones. In columns with typos or morphological
	variants of the same information, a simple string similarity is often
	enough to capture additional information. Similarity encoding outperforms
	a bag-of-n-grams representation of the input string, as well as 
	methods that encode high-cardinality categorical variables
	without capturing information in the strings representations
	\cite{cerda2018similarity}, such as \emph{target encoding}
	\cite{micci2001preprocessing} or \emph{hash encoding}
	\cite{weinberger2009feature}.
	
	
	A variety of string similarities can be considered for similarity
	encoding, but \cite{cerda2018similarity} found that a good performer
	was a similarity
	based on n-grams of consecutive characters. This n-gram similarity is based on
	splitting the two strings to compare in their character n-grams and 
	calculating the Jaccard coefficient between these two sets
	\cite{angell1983automatic}:
	\begin{equation}
	\label{eq:ngram_sim_dice}
	\text{sim}_\text{n-gram}(s_i^{}, s_j^{}) =
	J(\mathcal{G}(s_i^{}), \mathcal{G}(s_j^{})) =
	\frac{|\mathcal{G}(s_i^{}) \cap \mathcal{G}(s_j^{})|}
	{|\mathcal{G}(s_i^{}) \cup \mathcal{G}(s_j^{})|}
	\end{equation}
	where $\mathcal{G}(s)$ is the set of consecutive character n-grams
	for the string $s$.
	Beyond the use of string similarity, an important aspect of 
	similarity encoding is that it is a prototype method, using as prototypes a subset
	of the categories in the train set.
	
	\subsection{Related solutions for encoding string categories}

	\subsubsection{Bag of n-grams}
	
	A simple way to capture morphology in a string
	is to characterize it by the count of its character or word n-grams.
	This is sometimes called a \emph{bag-of-n-grams} characterization of strings.
	Such representation has been shown to be efficient for spelling correction
	\cite{angell1983automatic} or for named-entity recognition
	\cite{klein2003named}. Other vectorial representations, such as
	those created by neural networks, can also
	capture string similarities
	\cite{lu2019synergy}.
	
\begin{figure}
		\centering
		\includegraphics[trim={0 0 0 0},clip,height=.21\textheight]
		{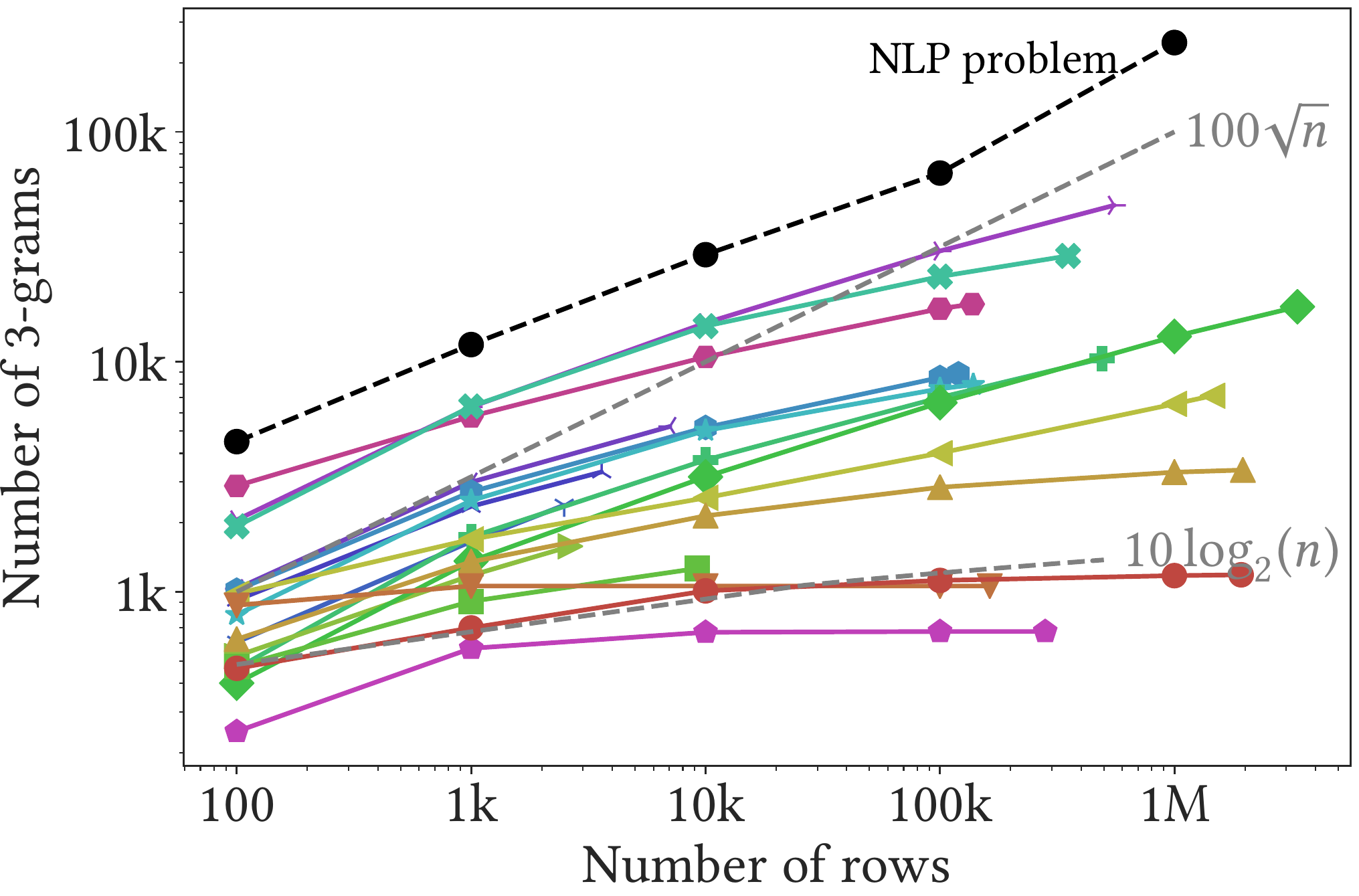}
		\caption{\textbf{Number of 3-gram
				versus number of samples} (colors as in
			\autoref{fig:cats_vs_nsamples}). The number of different n-grams tends to
			increase slower than in a typical NLP problem
			(Wikipedia articles in this case).}
		\label{fig:ngrams_vs_nsamples}
	\end{figure}

	For high-cardinality categorical variables, the
	number of different n-grams tends to increase with the number of samples.
	Yet, this number increases slower than in a typical NLP
	problem (see \autoref{fig:ngrams_vs_nsamples}).
	Indeed, categorical variables have less entropy than free text:
	they are usually repeated, often have subwords in common, and refer to
	a particular, more restrictive subject.
	
	Representing strings by character-level n-grams is related to
	vectorizing text by their tokens or words. Common practice uses 
	\emph{term-frequency} \emph{inverse-document-frequency}
	(\emph{tf-idf}) reweighting: dividing a token's count in a sample by its
	count in the whole document. Dimensionality reduction by a
	singular value decomposition (SVD) on this matrix leads to a simple topic
	extraction, latent semantic analysis (LSA)
	\cite{landauer1998introduction}. A related but more scalable solution
	for dimensionality reduction are random projections, which give low-dimensional 
	approximation of Euclidean distances
	\cite{johnson1984extensions,achlioptas2003database}.
	
	\subsubsection{Word embeddings}
	
	If the string entries are common words, an approach to represent them as vectors
	is to leverage word embeddings developed in 
	natural language processing
	\cite{pennington2014glove,mikolov2013efficient}.
	Euclidean similarity of these vectors captures related
	semantic meaning in words.
	Multiple words can be represented as a weighted sum of their vectors,
	or with more complex approaches \cite{arora2016simple}.
	To cater for out-of-vocabulary strings,
	FastText \cite{bojanowski2017enriching} 
	considers subword information of words,
	\emph{i.e.}, character-level n-grams. Hence, it can encode strings
	even in the presence of typos.
	Similarly, Bert \cite{devlin2018bert} uses also a composition of substrings to recover
	the encoding vector of a sentence.
	In both cases, word vectors computed on very large corpora are available for download.
	These have captured fine semantic links between words.
	However, to analyze a given database, the danger of such approach is that 
	the semantic of categories may differ from that in the pretrained model.
	These encodings do not 
	adapt to the information specific in the data at hand.
	Moreover, they cannot be trained directly on the categorical variables
	for two reasons: categories are typically short strings that do
	not embed enough context;
	and the number of samples in some datasets is not enough to properly
	train these models.
	
	\section{Scalable encoding of string categories}
	\label{sec:scalable_encoders}
	
	We now describe two novel approaches for categorical encoding of string
	variables.
	Both are based on the character-level structure of categories.
	The first approach, that we call \emph{min-hash encoding}, is
	inspired by the
	document indexation literature, and in particular the idea of
	locality-sensitive hashing (LSH) \cite{gionis1999similarity}. LSH
	gives a fast and stateless way to approximate
	the Jaccard coefficient between
	two strings \cite{broder1997resemblance}. The second approach is the
	\emph{Gamma-Poisson factorization} \cite{canny2004gap},
	a matrix factorization technique---originally used in the probabilistic topic
	modeling literature---that assumes a Poisson distribution on the
	n-gram counts of categories, with a Gamma prior on the activations.
	An online algorithm of the matrix factorization
	allows to scale the method with a linear complexity on the number of samples.
	Both approaches capture the morphological similarity
	of categories in a reduced dimensionality.
	
	\subsection{Min-hash encoding}
	
	\subsubsection{Background: min-hash}
	
	Locality-sensitive hashing (LSH) \cite{gionis1999similarity} has been
	extensively used for approximate nearest neighbor search for
	learning
	\cite{shrivastava2012fast,wang2017survey} or as an efficient way of finding similar
	objects (documents, pictures, etc.) \cite{chum2008near} in high-dimensional settings.
	One of the most famous functions
	in the LSH family is the min-hash function
	\cite{broder1997resemblance,broder2000min}, originally designed to retrieve
	similar documents in terms of the Jaccard coefficient of the word
	counts of documents (see \cite{leskovec2014mining}, chapter 3, for a
	primer). While min-hash is a classic tool for its collisions
	properties, as with nearest neighbors, we study it here as
	encoder for general machine-learning models.
	
	Let $\mathcal{X}^\star$ be a totally ordered set
	and $\pi$ a random permutation of its order.
	For any non-empty $\mathcal{X} \subseteq \mathcal{X}^\star$
	with finite cardinality,
	the min-hash function $Z(\mathcal{X})$ can be defined as:
	\begin{equation}
	Z(\mathcal{X}) \overset{\text{def}}{=}
	\min_{x \in \mathcal{X}} \pi(x)
	\end{equation}
	Note that $Z(\mathcal{X})$ can be also seen as a random variable.
	As shown in \cite{broder1997resemblance},
	for any $\mathcal{X}, \mathcal{Y} {\subseteq} \mathcal{X}^\star$,
	the min-hash function
	has the following property:
	\begin{equation}
	\mathbb{P}\left( Z(\mathcal{X}) {=} Z(\mathcal{Y}) \right)
	= \frac{|\mathcal{X} \cap \mathcal{Y}|}
	{|\mathcal{X} \cup \mathcal{Y}|}
	= J(\mathcal{X}, \mathcal{Y})
	\label{eq:collision_prob_0}
	\end{equation}
	Where $J$ is the Jaccard coefficient between the two sets.
	For a controlled approximation, several random
	permutations can be taken, which defines a min-hash signature.
	For $d$ permutations $\pi_j$ drawn \emph{i.i.d.}, 
	\autoref{eq:collision_prob_0} leads to:
	\begin{equation}
	\sum_{j = 1}^{d} \mathbbm{1}[Z_j(\mathcal{X}) = Z_j(\mathcal{Y})]
	\sim \mathcal{B}(d, J(\mathcal{X}, \mathcal{Y})).
	\label{eq:collision_prob}
	\end{equation}
	where $\mathcal{B}$ denotes the Binomial
	distribution. Dividing the above quantity by $d$
	thus gives a consistent estimate of the Jaccard
	coefficient $J(\mathcal{X}, \mathcal{Y})$\footnote{
		Variations of the min-hash algorithm,
		as the min-max hash \cite{ji2013min} can reduce the
		variance of the Jaccard similarity approximation.
}.
	
	
	Without loss of generality, we can consider the case of
	$\mathcal{X}^\star$ being equal to the real interval $[0, 1]$,
	so now for any $x \in [0, 1]$, $\pi_j(x) \sim \mathcal{U}(0, 1)$.
	\begin{proposition}{\bf Marginal distribution.}
		If $\pi(x) \sim \mathcal{U}(0, 1)$,
		and $\mathcal{X} {\subset} [0, 1]$ such that $|\mathcal{X}| {=} k$,
		then $Z(\mathcal{X}) \sim \text{Dir}(k,1)$.
		\begin{proof}
			It comes directly from considering that:\\
			$\mathbb{P}(Z(\mathcal{X}){\leq} z) =$
			$1 - \mathbb{P}(Z(\mathcal{X}) {>} z) =
			1 - \prod_{i=1}^{k} \mathbb{P}(\pi(x_i) > z) = 1 - (1-z)^k$.
		\end{proof}
		\label{prop:minhash_marginal}
	\end{proposition}
	Now that we know the distribution of the min-hash random variable, we will show how 
	each dimension of a min-hash signature maps inclusion of sets to simple inequalities.
	\begin{proposition}{\bf Inclusion.}
		Let $\mathcal{X}, \mathcal{Y} {\subset} [0, 1]$ such that $|\mathcal{X}| {=} k_x$
		and $|\mathcal{Y}| {=} k_y$.
		\begin{enumerate}[(i)]
			\item If $\mathcal{X} \subset \mathcal{Y}$, then $Z(\mathcal{Y}) \leq Z(\mathcal{X})$. \\
			\item $\mathbb{P}\bigl(Z(\mathcal{Y}) {\leq}
			Z(\mathcal{X})  \,\big|\,
			\mathcal{X} {\cap} \mathcal{Y} {=} \emptyset \bigr)
			=  \frac{k_y}{k_x + k_y}$
		\end{enumerate}
		\begin{proof}
			\textit{(i)} is trivial and \textit{(ii)} comes directly from
			Prop. \ref{prop:minhash_marginal}:
			\begin{eqnarray*}
				\lefteqn{\mathbb{P}\left(Z(\mathcal{Y}) {-} Z(\mathcal{X}) \leq 0
					\, | \, \mathcal{X} {\cap} \mathcal{Y}  {=} \emptyset \right)} \\
				& &= \int_{0}^{1}\int_{0}^{x} f_{Z(\mathcal{Y})}(y) f_{Z(\mathcal{X})}(y) \,  dy \, dx  \\ 
				& &= \int_{0}^{1} \left(1 - (1 - x)^{k_y} \right)
				k_x(1 - x)^{k_x - 1} dx
				= \frac{k_y}{k_x + k_y}
			\end{eqnarray*}
		\end{proof}
		\label{prop:minhash_subset}
	\end{proposition}
	At this point, we do not know anything about the case when
	$\mathcal{X} \not\subseteq \mathcal{Y}$,
	so for a fixed $Z(\mathcal{X})$, we can not ensure that any set with
	lower min-hash value has $\mathcal{X}$ as inclusion.
	The following theorem allows us to define regions in the vector space
	generated by the min-hash signature that, with high probability, are associated
	to inclusion rules.
	\begin{theorem}{\bf Identifiability of inclusion rules.}\\
		Let $\mathcal{X}, \mathcal{Y} \,{\subset}\, [0, 1]$ be two finite sets
		such that $|\mathcal{X}| {=} k_x$
		and $|\mathcal{Y}| {=} k_y$. $\forall \, \epsilon {>} 0$,
		if $d \,{\geq}\, \lceil {-} \log(\epsilon) / \log(1 {+} \frac{k_x}{k_y}) \rceil$, then:
		\begin{equation}
		\mathcal{X}\,{\not\subseteq}\,\mathcal{Y}
		\Rightarrow
		\mathbb{P}\left(\sum_{j = 1}^{d}
		\mathbbm{1}[Z_j(\mathcal{Y}) {\leq} Z_j(\mathcal{X})] = d \right) \leq \epsilon.
		\end{equation}
		\begin{proof}
			First, notice that:
			\begin{equation*}
			\mathcal{X} {\not\subseteq} \mathcal{Y} \iff
			\exists \, k \in \mathbb{N}, 0 \leq k < k_x \text{ such that }
			|\mathcal{X} {\cap} \mathcal{Y}| = k\\
			\end{equation*}
			Then, defining $\mathcal{Y'} \overset{\text{def}}{=} \mathcal{Y} \setminus  (\mathcal{X} {\cap} \mathcal{Y})$, with $|\mathcal{Y'}| = k_y - k$:
			\begin{align*}
			\mathbb{P}\left(Z(\mathcal{Y}) {\leq} Z(\mathcal{X})  \,|\,
			\mathcal{X} {\not\subseteq} \mathcal{Y} \right) &=
			\mathbb{P}\left(Z(\mathcal{Y'}) {\leq} Z(\mathcal{X})  \,|\,
			\mathcal{X} {\cap} \mathcal{Y'} = \emptyset \right)\\ 
			& = (k_y - k) / (k_x + k_y - k) \\
			& \leq k_y / (k_x + k_y )\\
			& = \mathbb{P}\left(Z(\mathcal{Y}) {\leq} Z(\mathcal{X})  \,|\,
			\mathcal{X} {\cap} \mathcal{Y} = \emptyset \right)
			\end{align*}	
			Finally:
			\begin{eqnarray*}
				\lefteqn{\mathbb{P}\left(\sum\nolimits_{j = 1}^{d}
					\mathbbm{1}[Z_j(\mathcal{Y}) {\leq} Z_j(\mathcal{X})] = d \,|\,
					\mathcal{X} {\not\subseteq} \mathcal{Y} \right)} \\
				&&= \mathbb{P}\left(Z(\mathcal{Y}) {\leq} Z(\mathcal{X})
				\, | \, \mathcal{X} {\not\subseteq} \mathcal{Y} \right) ^d\\ 
				&&\leq \mathbb{P}\left(Z(\mathcal{X}) {\leq} Z(\mathcal{Y})
				\, | \, \mathcal{X} {\cap} \mathcal{Y} {=} \emptyset \right) ^d
				= \left(\frac{k_y}{k_x + k_y}\right)^d
			\end{eqnarray*}
		\vspace{-10pt}
		\end{proof}
		\label{thm:inclusion_rules}	
	\end{theorem}
	Theorem \ref{thm:inclusion_rules} tells us that 
	taking enough random permutations ensures that
	when $\forall j, Z_j(\mathcal{Y}) {\leq} Z_j(\mathcal{X})$,
	the probability that $\mathcal{X}\,{\not\subseteq}\,\mathcal{Y}$ is small.
	This result is very important, as it shows a global property of the min-hash
	representation when using several random permutations, going beyond the well-known
	properties of collisions in the min-hash signature. \autoref{fig:employee_salaries_false_positives} in the Appendix confirms empirically
	the bound on the dimensionality $d$ and its logarithmic dependence on the
	desired false positive rate $\epsilon$.
	
	\subsubsection{The min-hash encoder}
	
	A practical way to build a computationally efficient implementation of
	min-hash is to use a hash function with different salt numbers instead of 
	random permutations. Indeed, hash functions can be built with suitable
	\emph{i.i.d.} random-process properties \cite{broder2000min}.
	Thus, the min-hash function can be constructed as follows:
	\begin{equation}
	Z_j(\mathcal{X}) = \min_{x \in \mathcal{X}} h_j(x),
	\end{equation}
	where $h_j$ is a hash function\footnote{
		Here we use a 32bit version of
		the MurmurHash3 function \cite{appleby2014murmurhash3}.}
	on $\mathcal{X}^\star$ with salt value $j$.
	
	For the specific problem of categorical data, we are interested in a fast
	approximation of $J(\mathcal{G}(s_i), \mathcal{G}(s_j))$, where
	$\mathcal{G}(s)$ is the set of all consecutive character n-grams for
	the string $s$.
	We define the min-hash encoder as:
	\begin{equation}
	\mathbf{x}^\text{min-hash} (s) \overset{\text{def}}{=}
	[Z_1(\mathcal{G}(s)), \dots,
	Z_d(\mathcal{G}(s))] \in \mathbb{R}^d.
	\end{equation}
	Considering the hash functions as random processes,
	\autoref{eq:collision_prob} implies that
	this encoder has the following property:
	\begin{equation}
	\frac{1}{d} \,  \mathbb{E}\left[
	\norm{\mathbf{x}^{\text{min-hash}}(s_i) -
		\mathbf{x}^{\text{min-hash}}(s_j)}_{\ell_0} \right] =
	J(\mathcal{G}(s_i), \mathcal{G}(s_j))
	\end{equation}
	Proposition \ref{prop:minhash_subset} tells us that the min-hash encoder
	transforms
	the inclusion relations of strings
	into an order relation in the feature space.
	This is especially relevant for learning tree-based models,
	as theorem \ref{thm:inclusion_rules}
	shows that by performing a reduced number of splits in the min-hash dimensions,
	the space can be divided between the elements that contain
	and do not contain a given substring $s$.
	\begin{figure}
		\centering
		\includegraphics[trim={0 0 0 0},clip,height=.24\textheight]
		{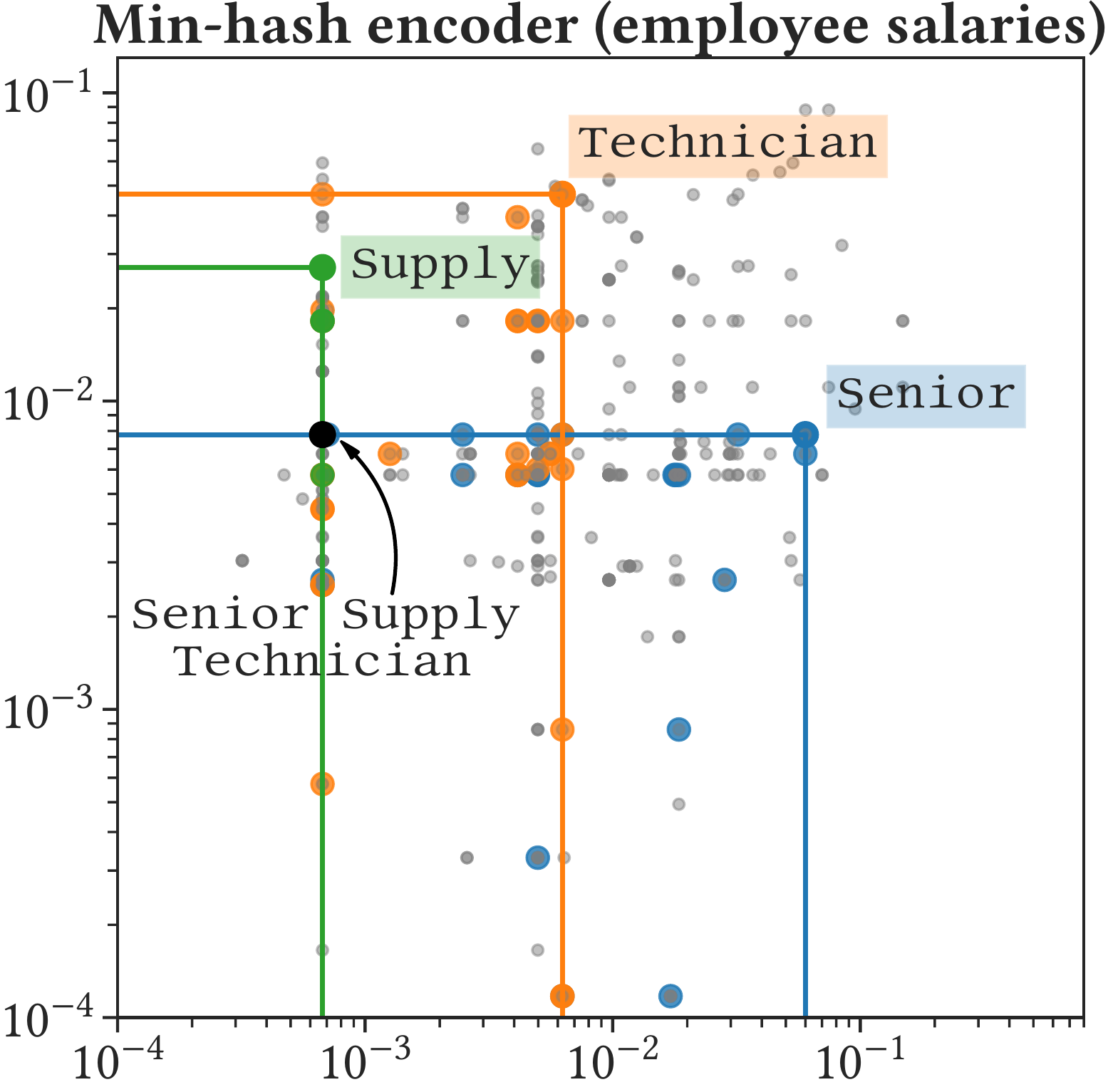}
		\caption{\textbf{The min-hash encoder transforms containment into
				inequality operations.}
			Color dots are categories that contain the corresponding colored
			substrings and gray dots are categories that do not contain any
			of them. The category \texttt{Senior Supply Technician} (black dot)
			is in the intersection of the three containment regions.}
		\label{fig:minhash_interpretation}
	\end{figure}
	As an example, \autoref{fig:minhash_interpretation} shows this global
	property of the min-hash encoder for the case of the employe salaries
	dataset with $d{=}2$.
	The substrings
	\texttt{Senior}, \texttt{Supply} and \texttt{Technician} are all included
	in the category \texttt{Senior Supply Technician}, and as consequence,
	the position for this category in the encoding space will be always in the
	intersection of the	bottom-left regions generated by its substrings.
	
	Finally, this encoder is specially suitable for very large scale settings,
	as it is very fast to compute and completely stateless. A stateless
	encoding is very useful for distributed computing: different
	workers can then process data simultaneously without 
	communication. Its drawback is that, as it relies on hashing, the encoding
	cannot easily be inverted and interpreted in terms of the original string
	entries.
	
	\subsection{Gamma-Poisson factorization}
	
	To facilitate interpretation, we now introduce an encoding approach that
	estimates a decomposition of the string entries in terms of a linear
	combination of latent categories.
	
	\subsubsection{Model}
	
	We use a generative model of strings from latent categories.
	For this, we rely on
	the Gamma-Poisson model \cite{canny2004gap}, a matrix
	factorization-technique well-suited to counting statistics.
	The idea was originally developed for finding low-dimensional representations,
	known as topics, of documents given their word count representation.
	As the string entries we consider are much shorter than text documents
	and can contain typos, we rely on their substring representation: we represent
	each observation by its count vector of
	character-level structure of n-grams.
	Each observation, a string entry described by its
	count vector $\mathbf{f} \in
	\mathbb{N}^{m}$, is modeled as a linear combination of $d$ unknown \textit{prototypes} or \textit{topics}, $\mathbf{\Lambda} \in
	\mathbb{R}^{d \times m}$:
	\begin{equation}
	\mathbf{f} \approx \mathbf{x} \, \mathbf{\Lambda},
	\end{equation}
	Here, $\mathbf{x} \in \mathbb{R}^d$ are the activations that decompose the
	observation $\mathbf{f}$ in the prototypes $\mathbf{\Lambda}$ in the count space.
	As we will see later, these prototypes can be seen as latent categories.
	
	Given a training dataset with $n$ samples, the model estimates the
	unknown prototypes $\mathbf{\Lambda}$ by factorizing the data's
	bag-of-n-grams representation
	$\mathbf{F} \in \mathbb{N}^{n \times m}$, where $m$ is the number of
	different n-grams in the data:
	\begin{equation}	
	\mathbf{F} \approx \mathbf{X} \, \mathbf{\Lambda},
	\quad \text{with } \mathbf{X} \in \mathbb{R}^{n \times d},
	\mathbf{\Lambda} \in \mathbb{R}^{d \times m}
	\end{equation}
	As $\mathbf{f}$ is a vector of counts, it is natural to consider a Poisson
	distribution for each of its elements:
	\begin{equation}
	p\bigl(f_j|(\mathbf{x} \, \mathbf{\Lambda})_j \bigr) =
	\frac{1}{f_j!} (\mathbf{x} \, \mathbf{\Lambda})_j^{f_j} e^{-(\mathbf{x} \,
		\mathbf{\Lambda})_j}, \quad j= 1, ..., m.
	\end{equation}
	For a prior on the elements of $\mathbf{x} \in \mathbb{R}^d$, we use a 
	Gamma distribution, as it is the conjugate prior of the
Poisson distribution, but also because it can foster a soft sparsity:
	\begin{equation}
	p(x_i) = \frac{x_i^{\alpha_i-1} e^{-x_i/\beta_i}}{\beta_i^{\alpha_i} \, \Gamma(\alpha_i)},
	\quad i= 1, ..., d,
	\end{equation}
	where $\boldsymbol{\alpha}$, $\boldsymbol{\beta}$ $\in \mathbb{R}^d$ are the shape and scale
	parameters of the Gamma distribution for each one of the $d$
	topics.
	
	\subsubsection{Estimation strategy}
	To fit the model to the input data, we maximize the log-likelihood
	of the model, written as:
	\begin{align}{}
	&\log \mathcal{L} &=& \quad \sum_{j=1}^{m} f_j \log ((\mathbf{x} \,
	\mathbf{\Lambda})_j) - (\mathbf{x} \, \mathbf{\Lambda})_j -
	\log (f_j!) \, + \nonumber \\ 
	&&& \quad \sum_{i=1}^{d} (\alpha_i{-}1) \log (x_i) -
	\frac{x_i}{\beta_i} - \alpha_i \log \beta_i - \log \Gamma(\alpha_i)
	\end{align}
	Maximizing the log-likelihood with respect to the parameters gives:
	\begin{align}
	\frac{\partial}{\partial \Lambda_{ij}} \log \mathcal{L} &=
	\frac{f_j}{(\mathbf{x} \, \mathbf{\Lambda})_j} x_i - x_i \\
	\frac{\partial}{\partial x_i} \log \mathcal{L} &=
	\sum_{j=1}^{m} \frac{f_j}{(\mathbf{x} \, \mathbf{\Lambda})_j} \Lambda_{ij} -
	\Lambda_{ij} + \frac{\alpha_i-1}{x_i} - \frac{1}{\beta_i}
	\end{align}
	As explained in \cite{canny2004gap}, these expressions are analogous to
	solving the following non-negative matrix factorization (NMF) with the
	generalized Kullback-Leibler divergence\footnote{
		In the sense of the NMF literature. See for instance
		\cite{lee2001algorithms}.} as loss:
	\begin{equation}
	\begin{pmatrix}
	\mathbf{F} \\
	\text{diag}(\boldsymbol{\beta})^{-1}
	\end{pmatrix} = \mathbf{X}
	\begin{pmatrix}
	\mathbf{\Lambda} \\
	\text{diag}(\boldsymbol{\alpha}) - I_d
	\end{pmatrix}
	\end{equation}
	In other words, the Gamma-Poisson model can be interpreted as a constrained
	non-negative matrix factorization in which the generalized
	Kullback-Leibler divergence is minimized between
	$\mathbf{F}$ and $\mathbf{X}\mathbf{\Lambda}$,
	subject to a Gamma prior in the distribution of the elements of $\mathbf{X}$.
	The Gamma prior induces sparsity in the activations $\mathbf{x}$
	of the model. 
	
	To solve the NMF problem above, \cite{lee2001algorithms} proposes the
	following recurrences:
	\begin{align}
	\label{ec:lambda_recursion}
	\Lambda_{ij} & \leftarrow
	\Lambda_{ij}
	\left(\sum_{\ell=1}^{n}
	\frac{f_{\ell j}}{(\mathbf{X} \mathbf{\Lambda})_{\ell j}}
	x_{\ell i} 
	\right)
	\left( \sum_{\ell =1}^{n} x_{\ell i} \right)^{-1} \\
	x_{\ell i} & \leftarrow
	x_{\ell i}
	\left( \sum_{j=1}^{m}
	\frac{f_{\ell j}}{(\mathbf{X} \mathbf{\Lambda})_{\ell j}}
	\Lambda_{ij} 
	+ \frac{\alpha_i - 1}{x_{\ell i}} \right)
	\left(\sum_{j=1}^{m} \Lambda_{ij} + \beta_i^{-1}\right)^{-1}
	\label{ec:lambda_recursion_2}
	\end{align}
	As $\mathbf{F}$ is a sparse matrix, the summations above
	only need to be computed on the non-zero elements of $\mathbf{F}$.
	This fact considerably decreases the computational cost of the algorithm.
	\begin{algorithm}[b!]
		\DontPrintSemicolon
		\SetAlgoLined
		\Input{$\mathbf{F} {\in} \mathbb{R}^{n \times m},
			\mathbf{\Lambda}^{(0)} {\in} \mathbb{R}^{d \times m},
			\boldsymbol{\alpha}, \boldsymbol{\beta} \in \mathbb{R}^{d},
			\rho,q,\eta,\epsilon$}
		\Output{$\mathbf{X} {\in} \mathbb{R}^{n \times d},
			\mathbf{\Lambda} {\in} \mathbb{R}^{d \times m}$}
		\vspace{.1cm}
		\While{ $\norm{\mathbf{\Lambda}^{(t)} - \mathbf{\Lambda}^{(t-1)}}_F > \eta$}{
			\vspace{.1cm}
			draw $\mathbf{f}_t^{}$ from the training set $\mathbf{F}$.\\
			\vspace{.15cm}
			\While{ $\norm{\mathbf{x}_t^{} -
					\mathbf{x}_t^{\text{old}}}_2 > \epsilon$}{
				\vspace{.1cm}
				$\mathbf{x}_t^{} \leftarrow
				\left[\mathbf{x}_t^{}
				\left(
				\frac{\mathbf{f}_t^{}}
				{\mathbf{x}_t^{} \mathbf{\Lambda}^{(t)}}
				\right) \mathbf{\Lambda}^{(t) \mathsf{T}} +
				\boldsymbol{\alpha}-1
				\right]. 
				\left[\mathbf{1} \, \mathbf{\Lambda}^{(t)\mathsf{T}} {+}
				\boldsymbol{\beta}_{}^{-1}
				\right]^{.-1}$
			}
			\vspace{.15cm}
			$\mathbf{\tilde{A}}_t \leftarrow
			\mathbf{\Lambda}^{(t)} .
			\left[
			\mathbf{x}_t^{\mathsf{T}}
			\left(
			\frac{\mathbf{f}_t}{\mathbf{x}_t^{} \mathbf{\Lambda}^{(t)}}
			\right)
			\right]$\\
			$\mathbf{\tilde{B}}_t \leftarrow \mathbf{x}_{t}^{\mathsf{T}} \mathbf{1}$\\
			\vspace{.2cm}
			\If{$t \equiv 0 \bmod q$, \tcp*{Every $q$ iterations}}{
				\vspace{.1cm}
				$\mathbf{A}^{(t)} \leftarrow \rho \, \mathbf{A}^{(t-q)} +
				\sum_{s=t-q+1}^{t} \mathbf{\tilde{A}}^{(s)}$\\
				$\mathbf{B}^{(t)} \leftarrow \rho \, \mathbf{B}^{(t-q)} +
				\sum_{s=t-q+1}^{t} \mathbf{\tilde{B}}^{(s)}$\\
				$\mathbf{\Lambda}^{(t)} \leftarrow \mathbf{A}^{(t)} ./ \,
				\mathbf{B}^{(t)}$\\
			}
			\vspace{.1cm}
			$t \leftarrow t + 1$ \\
		}
		\caption{Online Gamma-Poisson factorization}
		\label{alg:online_GP}
	\end{algorithm}
	Following \cite{lefevre2011online}, we present an online (or streaming) version of the
	Gamma-Poisson solver (\autoref{alg:online_GP}).
	The algorithm exploits the fact that in the recursion
	for $\mathbf{\Lambda}$ (eq.\,\ref{ec:lambda_recursion} and \ref{ec:lambda_recursion_2}), the summations are done with respect to the training
	samples. Instead of computing the numerator and denominator in the entire
	training set at each update, one can update them only with mini-batches
	of data, which considerably decreases the memory usage and time of the
	computations. 
	
	For better computational performance,
	we adapt the implementation of this solver to the specificities
	of our problem---factorizing substring counts across entries of a
	categorical variable.
	In particular, we take advantage of the repeated entries
	by saving a dictionary of the activations for each category in the
	convergence of the previous mini-batches (\autoref{alg:online_GP}, line 4)
	and use them as an initial guess
	for the same category in a future mini-batch.
	This is a warm restart and is especially important in the case of
	categorical variables because for most datasets, the number of unique
	categories is much lower than the number of samples.
	
	We also set the hyper-parameters of the algorithm and its
	initialization for optimal convergence.
	For $\rho$, the discount factor for the previous
	iterations of the topic matrix $\mathbf{\Lambda}^{(t)}$
	(\autoref{alg:online_GP}, line 9-10).
	choosing
	$\rho{=}0.95$ gives good convergence speed while avoiding
	instabilities
	(\autoref{fig:benchmark_rho_gamma-poisson} in the Appendix).
	With respect to the initialization of the topic matrix
	$\mathbf{\Lambda}^{(0)}$,
	a good option is to choose the centroids of a k-means clustering
	(\autoref{fig:benchmark_init_gamma-poisson}) in a
	hashed version\footnote{We use the ``hashing trick''
	    \cite{weinberger2009feature} to construct a feature matrix without
	    building a full vocabulary, as this avoids a pass on the data
	    and creates a low-dimension representation.}
	of the n-gram count matrix $\mathbf{F}$
	and then use as initializations the nearest neighbor observations
	in the original n-gram space.
	In the case of a streaming setting, the same approach can be used in a subset of
	the data.
		
	\subsubsection{Inferring feature names}
	\label{sec:inferring_feature_names}

	An encoding strategy where each dimension can be understood
	by humans facilitates the interpretation of the full statistical
	analysis.
	A straightforward strategy for interpretation of the Gamma Poisson
	encoder is
	to describe each encoding dimension by features of the string entries
	that it captures. 
	For this, one alternative is to track
	the feature maps corresponding to each input category, and assign labels
	based on the input categories that activate the most in a given dimensionality.
	Another option is to apply the same strategy,
	but for substrings, such as words
	contained in the input categories.
	In the experiments, we follow the second approach as
	a lot of datasets are composed of entries with overlap,
	hence individual words carry more information for interpretability
	than the entire strings.
	
	This method is expected to work well
	if the encodings are sparse and composed only of non-negative values with a
	meaningful magnitude. The Gamma-Poisson factorization model
	ensures these properties.
	
	\begin{table*}[t!]
		\caption{\textbf{Non-curated datasets.} Description for the corresponding
			high-cardinality categorical variable.}
		\label{tab:datasets}       
		\centering
		\setlength\tabcolsep{5pt}
		\rowcolors{2}{black!5}{white}
		\begin{tabular}
			{L{.157\linewidth}  R{.066\linewidth} R{.079\linewidth} R{.106\linewidth}
				R{.073\linewidth} R{.105\linewidth} L{.2\linewidth}}
			\hline\noalign{\smallskip}
			\textbf{Dataset} &  \textbf{\#samples} & 
			\textbf{\#categories} & \textbf{\#categories per 1000 samples} &		
			\textbf{Gini coefficient} &
			\textbf{Mean category length (\#chars)} &
			\textbf{Source of high cardinality} \\
			\noalign{\smallskip}\hline\noalign{\smallskip}
			\input{table_df_datasets.txt}
			\hline
		\end{tabular}
	\end{table*}

	\section{Experimental study}
	\label{sec:experiments}
	
	We now study experimentally different encoding methods
	in terms of interpretability and supervised-learning performance.
	For this purpose, we use three different types of data:
	simulated categorical data, and real data with curated and non-curated
	categorical entries.
	
	We benchmark the following strategies:
	one-hot, tf-idf, fastText \cite{mikolov2018advances},
	Bert \cite{devlin2018bert},
	similarity encoding \cite{cerda2018similarity},
	the Gamma-Poisson factorization\footnote{
		Default parameter values are listed in
		\autoref{tab:parameters_gamma_poisson}},
	and min-hash encoding.
	For all the strategies based on a n-gram representation,
	we use the set of 2-4 character grams\footnote{
		In addition to the word as tokens,
		pretrained versions of fastText also use the set of 3-6 character
		n-grams.}.
	For a fair comparison across encoding strategies, we use the same
	dimensionality $d$ in all approaches.
	To set the dimensionality of one-hot encoding, tf-idf and fastText,
	we used a truncated SVD (implemented efficiently following
	\cite{halko2011finding}). Note that dimensionality reduction improves
	one-hot encoding with tree-based learners for data with rare categories
	\cite{cerda2018similarity}.
	For similarity encoding, we select prototypes 
	with a \textit{k-means} strategy,
	as it gives slightly better prediction results than the
	\textit{most frequent categories}\footnote{An implementation of these strategies can be found on \url{https://dirty-cat.github.io} \cite{cerda2018similarity}.
		We do not test the \textit{random projections} strategy for
		similarity encoding as it is not scalable. }.

	\subsection{Real-life datasets with string categories}
	
	\subsubsection{Datasets with high-cardinality categories}
	
	In order to evaluate the different encoding strategies, we collected
	17 real-world datasets containing a prediction task and at least one
	relevant high-cardinality categorical variable as feature\footnote{
		If a dataset has more than one categorical variable,
		only one selected variable was encoded with the proposed approaches,
		while the rest of them were one-hot encoded.}.
	\autoref{tab:datasets} shows a quick description of the datasets and
	the corresponding categorical variables (see Appendix
	\ref{sec:dataset_description} for a description of datasets and
	the related learning tasks).
	\autoref{tab:datasets} also details the source of high-cardinality for
	the datasets: \textit{multi-label}, \textit{typos},
	\textit{description} and \textit{multi-language}.
	We call \textit{multi-label} the situation when
	a single column contains multiple information shared by several entries, \emph{e.g.}, \texttt{supply technician}, where \texttt{supply} denotes the type of activity, and \texttt{technician} denotes the rank of the employee
	(as opposed, \emph{e.g.}, to \texttt{manager}). \textit{Typos} refers
	to entries having small morphological variations, as \texttt{midwest}
	and \texttt{mid-west}. \textit{Description} refers to categorical entries that
	are composed of a short free-text description. These are close to a
	typical NLP problem,
	although constrained to a very particular subject,
	so they tend to contain very recurrent informative words and
	near-duplicate entries.
	Finally, \textit{multi-language} are datasets in which the categorical
	variable contains more that one language across the different entries.
	
	\subsubsection{Datasets with curated strings}
	
	We also evaluate encoders when the categorical variables
	have already been curated: often, entries are standardized to
	create 
	low-cardinality categorical variables. For this, we collected seven
	of such datasets
	(see Appendix \ref{subsubsec:description_curated_datasets}).
        On these datasets we study the robustness of the
	n-gram based approaches to situations where there is no a priori need to reduce
	the dimensionality of the problem.
	
	\subsection{Recovering latent categories}
	\label{sec:recovering_latent_categories}
	
	\subsubsection{Recovery on simulated data}
	
	\autoref{tab:datasets} shows that the most common
	scenario for high-cardinality string variables are
	\textit{multi-label} categories. The second most common problem is the
	presence of \textit{typos} (or any source of morphological variation of the same
	idea). To analyze these two cases in a controlled setting, we create two
	simulated sets of categorical variables. \autoref{tab:simulated_data_examples}
	shows examples of generated categories, taking as a base 8
	ground-truth categories of animals (details
	in Appendix \ref{sec:synthetic_data}).
	
	To measure the ability of an encoder to recover a feature matrix close to a
	one-hot encoding matrix of ground-truth categories in these simulated settings,
	we use the Normalized Mutual Information (NMI) as metric.
	Given two random variables $X_1$ and $X_2$, the NMI is defined as:
	\begin{equation}
	\text{NMI} = 2 \, \frac{I(X_1; X_2)}{H(X_1) + H(X_2)}
	\end{equation}
	Where $I(\cdot \, ; \cdot)$ is the mutual information and $H(\cdot)$
	the entropy.
	To apply this metric to the feature matrix $\mathbf{X}$ generated by the
	encoding of all ground truth
	categories, we consider $\mathbf{X}$ --after
	rescaling with an $\ell_1$ normalization of the rows--
	as a two dimensional probability distribution.
	For encoders that produce feature matrices with negative values,
	we take the element-wise absolute value of $\mathbf{X}$.
	The NMI is a classic measure of correspondences between clustering
	results \cite{vinh2010information}. Beyond its information-theoretical
	interpretation, an appealing property is that it is invariant to
	order permutations.
	The NMI of any permutation of the identity matrix is equal to 1 and the
	NMI of any constant matrix is equal to 0. Thus, the NMI in this case
	is interpreted as a recovering metric of a one-hot encoded matrix of latent,
	ground truth, categories.

	\autoref{tab:nmi_encoders} shows the NMI for both simulated
	datasets. The Gamma-Poisson factorization obtains the highest values in
	both multi-label and typos settings and for different dimensionalities of
	the encoders. The best recovery is obtained when the
	dimensionality of the encoder is equal to the number of ground-truth
	categories, \textit{i.e.}, $d{=}8$.

	\begin{table}[t!]
		\caption{Examples of simulated categorical variables.}
		\label{tab:simulated_data_examples}       
		\centering
		\footnotesize
		\begin{tabular}{l l}
			\toprule
			Type & Example categories \\
			\midrule
			Ground truth & \texttt{chicken}; \texttt{eagle}; \texttt{giraffe};
			\texttt{horse}; \texttt{leopard}; \\
			& \texttt{lion}; \texttt{tiger}; \texttt{turtle}. \\
			\rowcolor{black!5}
			Multi-label  & \texttt{lion chicken}; \texttt{horse eagle lion}. \\
			Typos (10\%) & \texttt{itger}; \texttt{tiuger}; \texttt{tgier};
			\texttt{tiegr}; \texttt{tigre}; \texttt{ttiger}. \\
			\hline
		\end{tabular}
	\end{table}

	\begin{table}[t!]
		\caption{\textbf{Recovery of categories for
simulations}: Normalized mutual
			information (NMI) for different encoders.}
		\label{tab:nmi_encoders}       
		\centering
		\footnotesize
		\rowcolors{3}{white}{black!5}
		\begin{tabular}{l|ccc|ccc}
			\toprule
			\multirow{2}{*}{\textbf{Encoder}} &
			\multicolumn{3}{c|}{\textbf{Multi-label}} &
			\multicolumn{3}{c}{\textbf{Typos}} \\
			& $d$=6 & $\mathbf{d}$\textbf{=8} & $d$=10 & $d$=6 &
			$\mathbf{d}$\textbf{=8} & $d$=10 \\
			\midrule
			Tf-idf + SVD       & 0.16 & 0.18 & 0.17 & 0.17 & 0.16 & 0.16 \\
			FastText + SVD     & 0.08 & 0.09 & 0.09 & 0.07 & 0.08 & 0.08 \\
			Bert + SVD     	   & 0.03 & 0.03 & 0.03 & 0.05 & 0.06 & 0.06 \\
			Similarity Encoder & 0.32 & 0.25 & 0.24 & 0.72 &
0.82 & \textbf{0.80} \\
			Min-hash Encoder   & 0.14 & 0.15 & 0.13 & 0.14 & 0.15 & 0.13 \\
			Gamma-Poisson      & \textbf{0.76} & \textbf{0.82} & \textbf{0.79} &
			\textbf{0.77} & \textbf{0.83} & \textbf{0.80} \\
			\hline
		\end{tabular}
	\end{table}
	

	\subsubsection{Results for real curated data}
	
	\begin{table}[t!]
		\caption{\textbf{Recovering true categories for curated
		entries.} NMI for different encoders ($d$=30) -- Appendix
 \ref{sec:additional_figures} gives results for different dimensions.}
		\label{tab:nmi_encoders_clean}       
		\centering
		\footnotesize
		\setlength\tabcolsep{2.0pt}
		\rowcolors{3}{white}{black!5}
		\begin{tabular}{lccccc}
			\toprule
			\textbf{Dataset} & Gamma   & Similarity & Tf-idf 	& FastText &Bert\\
			(cardinality) 	& Poisson & Encoding   &
			+ SVD 			& + SVD & + SVD \\
			\midrule
			{\bf Adult} (15) 			 & \textbf{0.75} & 0.71 & 0.54 & 0.19 & 0.07 \\
			\textbf{Cacao Flavors} (100) & \textbf{0.51} & 0.30 & 0.28 & 0.07  &  0.04 \\
			{\bf California Housing} (5) & 0.46 & 0.51 & \textbf{0.56}  & 0.20& 0.05 \\
			{\bf Dating Profiles} (19) 	 & \textbf{0.52} & 0.24 & 0.25 & 0.12 & 0.05 \\
			{\bf House Prices} (15) 	 & \textbf{0.83} & 0.25 & 0.32 & 0.11& 0.05	\\
			\textbf{House Sales} (70) 	 & \textbf{0.42} & 0.04 & 0.18 & 0.06& 0.02	\\
			\textbf{Intrusion Detection} (66) & 0.34 & \textbf{0.58} & 0.46 & 0.11& 0.05 \\
			\hline
		\end{tabular}
	\end{table}

	For curated data, the cardinality is usually low.
	We nevertheless perform the encoding using a default choice of
	$d=30$,	to gauge how well turn-key
	generic encoding represent these curated strings.
	\autoref{tab:nmi_encoders_clean} shows the NMI values for the different
	curated datasets,
	measuring how much the generated encoding resembles a
	one-hot encoding on the curated categories. 
	Despite the fact that it is used with a dimensionality larger than the
	cardinality of the curated category,
	Gamma-Poisson factorization has the highest recovery performance in 5 out of 7
	datasets\footnote{\autoref{tab:nmi_encoders_clean_truedim} in the Appendix
		show the same analysis but for $d{=}|C|$, the actual cardinality of the
		categorical variable.
		In this setting, the Gamma-Poisson gives much higher recovery
		results.}.

	\begin{figure}[t!]
		\begin{subfigure}[t]{.46\textwidth}
			\includegraphics[trim={7.5cm -1cm 12cm 11.7cm},clip,height=.08\textheight]
			{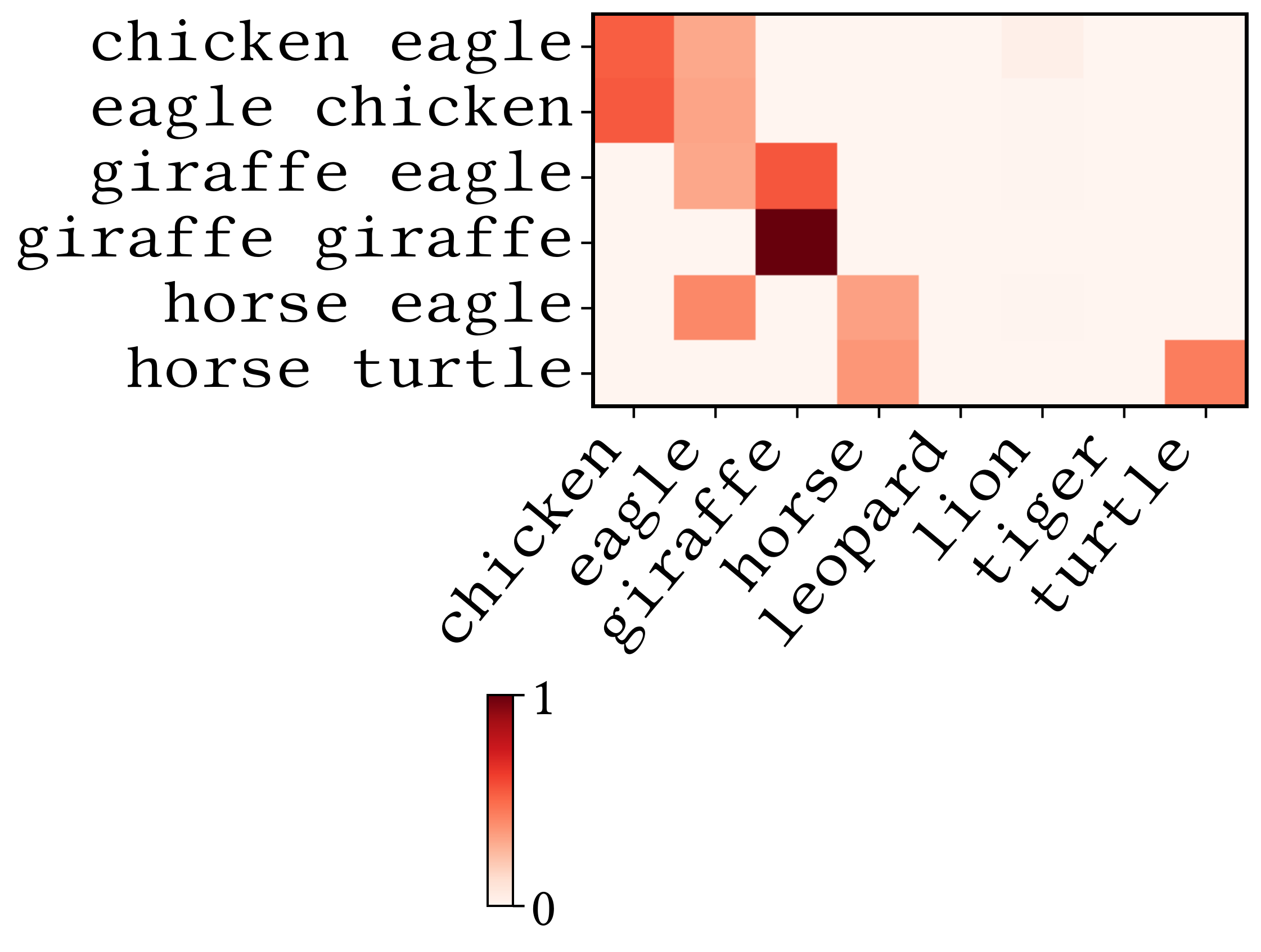}%
			\includegraphics[trim={0cm .3cm 0cm 0cm},clip,height=0.36\textheight]
			{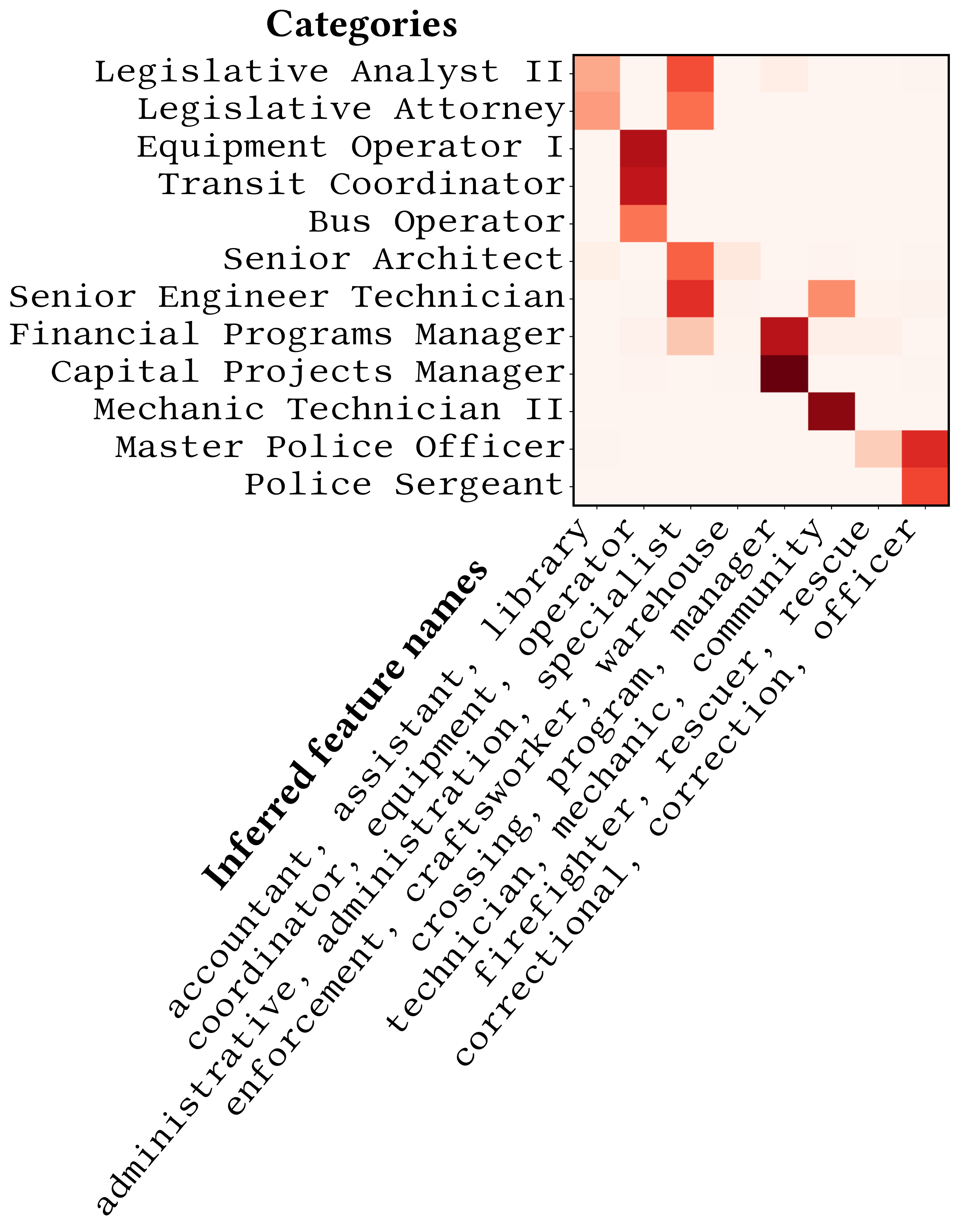}
			\caption{Employee Salaries dataset (Occupation)}
		\end{subfigure}%
		\vspace{.2cm}
		\begin{subfigure}[t]{.21\textwidth}
			\includegraphics[trim={0 0cm 0cm 0cm},clip,height=.105\textheight]
			{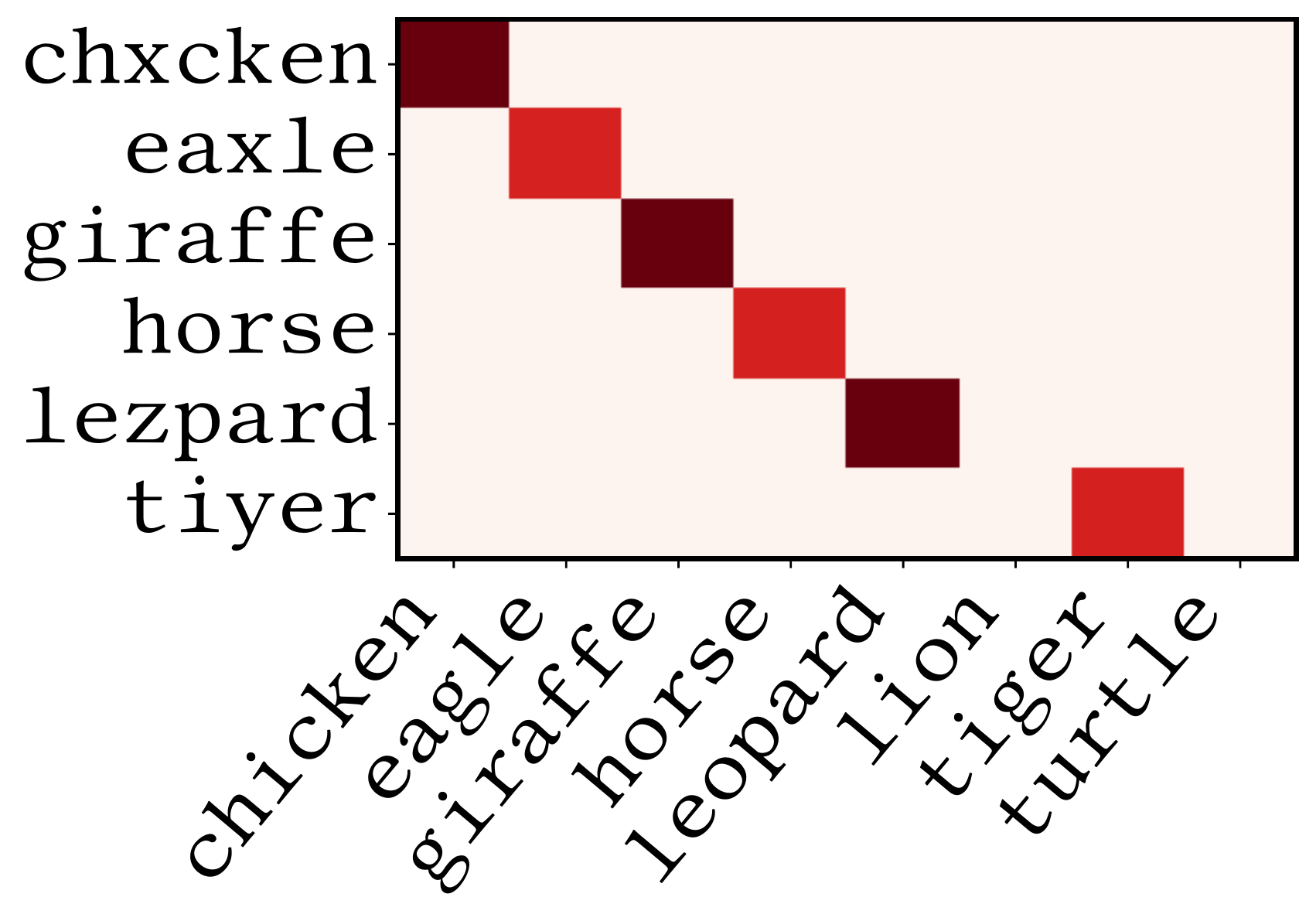}
			\caption{Simulated typos}
		\end{subfigure}%
		\begin{subfigure}[t]{.275\textwidth}
			\hfill
			\includegraphics[trim={0 5.2cm 0cm 0cm},clip,height=.105\textheight]
			{gamma_poisson_loadings_simulation_multilabel_transposed}
			\caption{Simulated multi-label categories}
		\end{subfigure}
		\caption{\textbf{Gamma-Poisson factorization gives positive and
				sparse representations that are easily interpretable.}
			Examples of encoding vectors (d=8) for a real dataset (a) and for simulated
			data (b and c) obtained with a Gamma-Poisson factorization.
			The $x$-axis shows the activations for each dimension with their
			respective inferred feature names. \autoref{fig:loadings_encoders_simulation_appendix}
			in the Appendix shows that other encoders fail to
			give such an easily-understandable picture.
		}
		\label{fig:loadings_encoders_simulation} 
	\end{figure}
	
	These experiments show that Gamma-Poisson factorization recovers
	well latent categories. To validate this intuition,
	\autoref{fig:loadings_encoders_simulation} shows such encodings in the
	case of the simulated data as well as the real-world non-curated
	Employees Salaries dataset. It confirms that the encodings can be
	interpreted as loadings on discovered categories that match
	the inferred feature names.	
	
	\subsection{Encoding for supervised learning}
	
	We now study the encoders for statistical analysis by 
	measuring prediction accuracy in supervised-learning tasks.
	
	\subsubsection{Experiment settings}
	
	We use gradient boosted trees, as implemented in
	XGBoost
	\cite{chen2016xgboost}.
	Note that trees can be implemented on categorical variables\footnote{
		XGBoost does not support categorical features. 
		The recommended option is to use one-hot encoding (\url{https://xgboost.readthedocs.io}).}. However,
	this encounter the same problems as one-hot encoding: the number of
	comparisons grows with the number of categories. Hence, the best trees
	approaches for categorical data use target encoding to impose an order on
	categories
	\cite{prokhorenkova2018catboost}. We also investigated other
	supervised-learning approaches: linear models, multilayer perceptron, and kernel
	machines with RBF and polynomial kernels. However, even with significant
	hyper-parameter tuning, they under-performed XGBoost on our tabular
	datasets (\autoref{fig:nemenyi_classifiers} in the Appendix).
	The good performance of gradient-boosted trees is consistent
	with previous reports of systematic benchmarks \cite{olson2017data}.
	
	Depending on the dataset, the learning task can be either
	\emph{regression}, \emph{binary} or \emph{multiclass} classification\footnote{
		We use different scores to evaluate the performance of the
		corresponding supervised
		learning problem: the $R^2$ score for regression;
		average precision for binary
		classif.;
		and accuracy for multiclass classif.}.
	As datasets get different prediction scores, we visualize 
	encoders' performance with prediction results scaled
	in a \emph{relative score}. It is a dataset-specific scaling of the
	original score, in order to bring performance across datasets in the same
	range. In other words, for a given dataset $i$:
	\begin{equation}
	\text{relative score}^i_j = 100
	\frac{\text{score}^i_j -
		\min_{j}{\text{score}^i_j}}
	{\max_{j}{\text{score}^i_j} -
		\min_{j}{\text{score}^i_j}} 
	\label{eq:relative_score}
	\end{equation}
	where $\text{score}^i_j$ is the the prediction score for the dataset $i$
	with the configuration $j {\in} \mathcal{J}$, the set of all trained
	models---in terms of dimensionality, type of encoder and 
	cross-validation split. The relative score is figure-specific and
	is only intended to be used as a visual comparison of classifiers' performance
	across multiple datasets. A higher relative score means better results.
	
	For a proper statistical comparison of encoders,
	we use a ranking test across multiple datasets
	\cite{demvsar2006statistical}. Note that in such a test each dataset
	amounts to a single sample, and not the cross-validation splits which are 
	not mutually independent.
	To do so, for a particular dataset, encoders were ranked according to the median
	score value over cross-validation splits.
	At the end, a Friedman test \cite{friedman1937use} is used to determine
	if all encoders, for a fixed dimensionality $d$, come
	from the same distribution.
	If the null hypothesis is rejected, we use a
	Nemenyi post-hoc test \cite{nemenyi1962distribution}
	to verify whether the difference in performance across
	pairs of encoders is significant.
	
	To do pairwise comparison between two encoders, we use a pairwise
Wilcoxon signed rank test. The corresponding p-values rejects the null
	hypothesis that the two encoders are equally performing across different
	datasets.
	
	\subsubsection{Prediction with non-curated data}

	We now describe the results of several prediction benchmarks
	with the 17 non-curated datasets.
	
	First, note that one-hot, tf-idf and fastText are naturally high-dimensional
	encoders, so a dimensionality reduction technique needs to be applied in order
	to compare the different methodologies---also, without this reduction,
	the benchmark will be unfeasible given the long computational
	times of gradient boosting. Moreover, dimensionality reduction helps to
	improve prediction (see \cite{cerda2018similarity}) with tree-based methods.
	To approximate Euclidean distances, SVD is optimal.
	However, it has a cost of $n d \min (n,d)$.
	Using Gaussian random projections \cite{rahimi2008random} is
	appealing, as can lead to stateless encoders that requires no
	fit.
	\autoref{tab:rand_proj_vs_svd} compares
	the prediction performance of both strategies.
	For tf-idf and fastText, the SVD is significantly superior to random
	projections.
	On the contrary, there is no statistical difference for one-hot, even
	though the performance is slightly superior for the SVD
	(p-value equal to 0.492).
	Given these results, we use SVD for all further benchmarks.

	\begin{table}[t!]
		\centering
		\caption{\textbf{Comparing SVD and Gaussian random
projection as a dimensionality reduction} Wilcoxon test p-values for
different encoders. Prediction performance with SVD is significantly superior for
			tf-idf, FastText and Bert.}
		\label{tab:rand_proj_vs_svd}
		\rowcolors{2}{black!5}{white}
		\begin{tabular}{lc}
			\toprule
			\textbf{Encoder} & \textbf{SVD v/s Random projection} (p-value) \\
			\midrule
			Tf-idf 		     & 					   \textbf{0.001} \\
			FastText         &					   \textbf{0.006} \\
			Bert        	 &					   \textbf{0.001} \\
			One-hot          &								0.717 \\
			\hline
		\end{tabular}
	\end{table}

	\begin{figure}
		\begin{subfigure}[b]{0.5\textwidth}
			\centering
			\includegraphics[trim={3.2cm 11.2cm 0cm 0cm},clip,width=.95\textwidth]
			{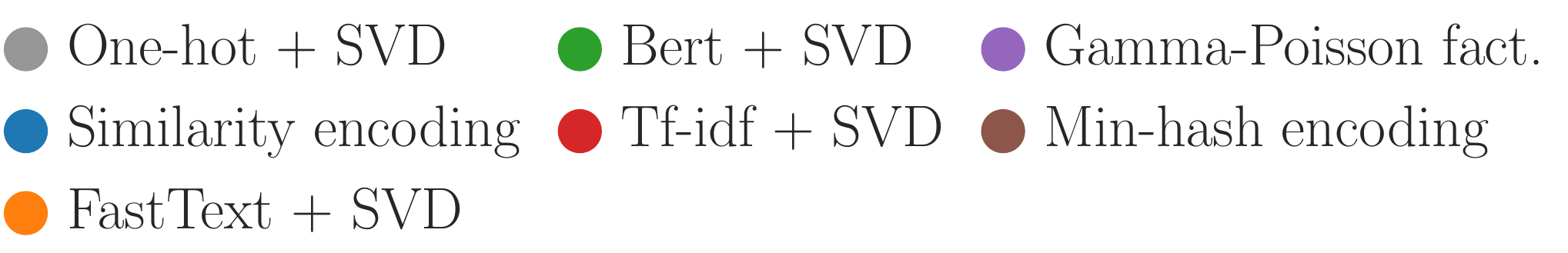}%
			\vspace{.1cm}
			\includegraphics[trim={0cm 0 6.0cm 4.05cm},clip,width=.95\textwidth]
			{datasets_encoders_across_dims_XGB}
			\label{fig:datasets_encoders_across_dims} 
		\end{subfigure}%
		\caption{\textbf{Encoding with subword information performs
				significantly better than one-hot.} Classifier: XGBoost.
			Comparison of encoders in terms of a relative score
			(the prediction score on the particular dataset,
			rescaled with respect to the global maximum and
			minimum score values across dimensions).
			Color dots indicate the scores for each cross-validation
			fold, black dots the median score across folds for a
			dataset, the black line indicates the
			median score and the box gives the interquartile range.
		}
		\label{fig:encoders_benchmark_non_curated}
	\end{figure}

	\autoref{fig:encoders_benchmark_non_curated} compares encoders in terms of
	the relative score of \autoref{eq:relative_score}. 
	All n-gram based encoders clearly improve
	upon one-hot encoding, at both dimensions ($d$ equal to 30 and 100).
	Min-hash gives a slightly better prediction
	performance across datasets, despite of being the only
	method that does not require a data fit step.
	The Nemenyi ranking test confirms the visual impression: n-gram-based methods are superior to one-hot encoding; and
	the min-hash encoder has the best average ranking value for both
	dimensionalities, although the difference in prediction with respect to
	the other n-gram based methods is not statistically significant.
	
\begin{figure}[t!]
		\begin{minipage}[c]{0.228\textwidth}
			\caption{\textbf{FastText prediction performance drops languages
					other than English.}
				Relative prediction scores with pretrained fastText
				vectors in different languages.
				The dimensionality was set with an SVD.
				A pairwise Wilcoxon signed rank tests give the following
				p-values:\\
				English-French $p$=0.056,
				French-Hungarian $p$=0.149,
				English-Hungarian $p$=0.019.
			}
		\vspace{-.4cm}
			\label{fig:fasttext_language_comparison}
		\end{minipage}
		\hfill
		\begin{minipage}[c]{0.25\textwidth}
			\includegraphics[trim={.5cm 0cm 0cm 1.0},clip,width=1\textwidth]
			{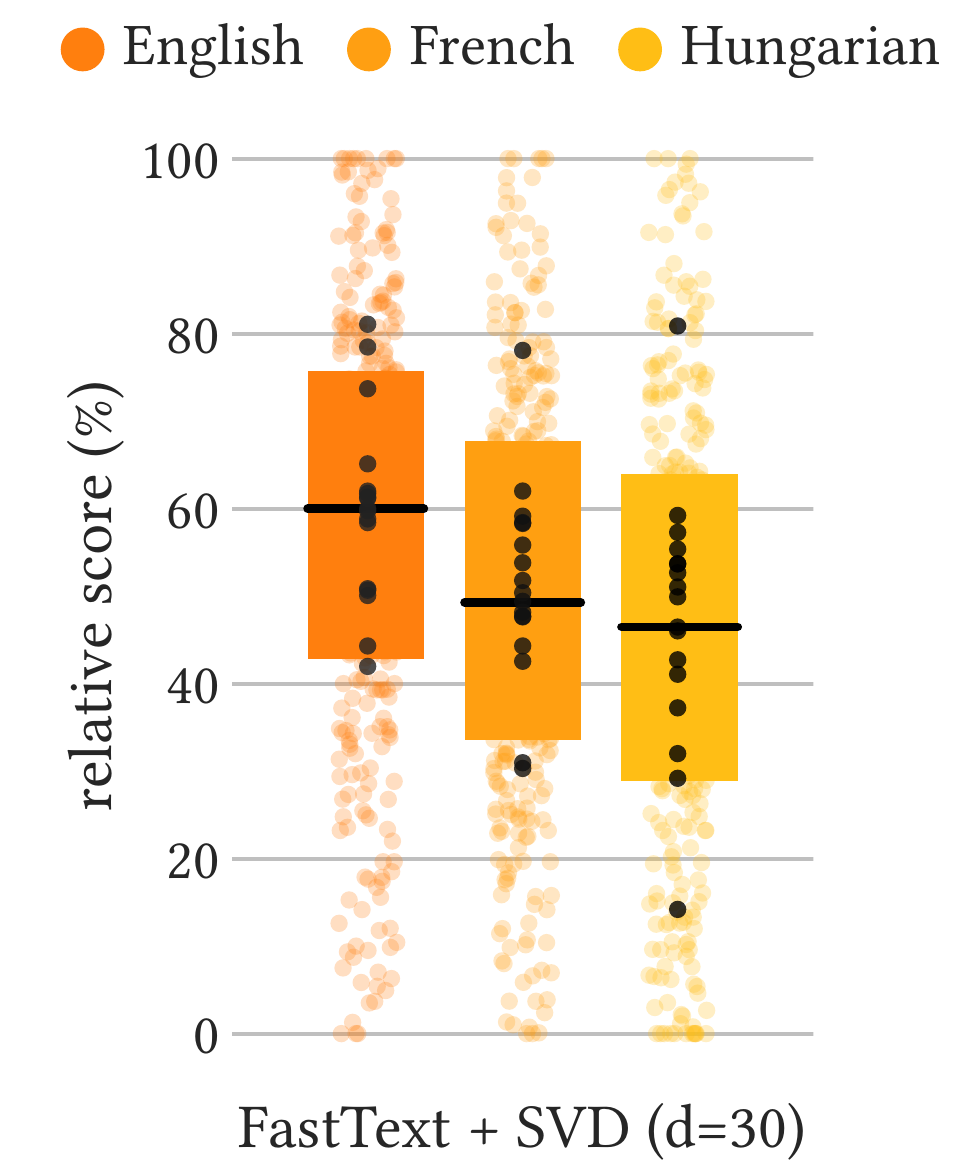}
		\end{minipage}
	\end{figure}

	While we seek \emph{generic} encoding approaches, using
	precomputed embeddings
	requires the choice of a language. As 15 out of 17 datasets are
	fully in English, the benchmarks above use English embeddings for
	fastText.
	\autoref{fig:fasttext_language_comparison},
	studies the importance of this choice, 
	comparing the prediction results for fastText
	in different languages (English, French and Hungarian).
	Not choosing English leads to a sizeable drop in prediction
	accuracy, which gets bigger for languages more distant (such as
	Hungarian).
	This shows that the natural language semantics of fastText indeed
	are important to explain its good prediction performance.
	A good encoding not only needs to represent
	the data in a low dimension, but also to capture the similarities
	between the different entries.

	\begin{figure}
		\begin{minipage}[c]{0.222\textwidth}
			\caption{\textbf{All encoders perform well
					for low-cardinality datasets.} Classifier: XGBoost. 
				The score is relative to the
				best and worse prediction across datasets
				(\autoref{eq:relative_score}).
				Color dots indicate the scores for each cross-validation
				fold, black dots the median across folds,
				the black line indicates the
				median across datasets and the box gives
				the interquartile range. Differences are not significant.}
			\vspace{-.2cm}
			\label{fig:encoders_benchmark_curated} 
		\end{minipage}
		\hfill
		\begin{minipage}[c]{0.25\textwidth}
			\includegraphics[trim={3.25cm 14.3cm 0cm 0cm},clip,width=1.0\textwidth]
			{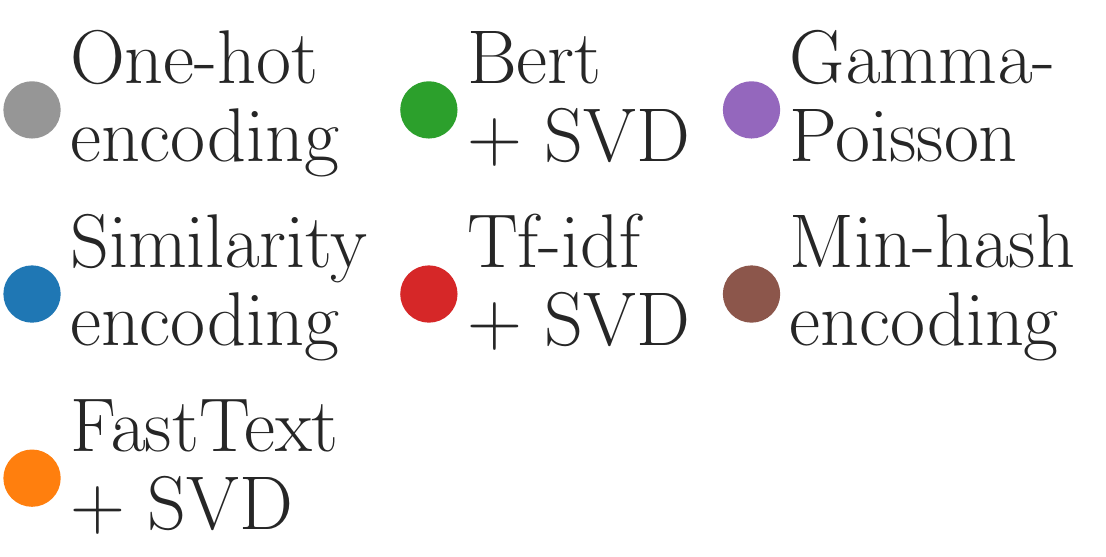}%
			\vspace{.1cm}
			\includegraphics[trim={0cm .1cm 5.6cm 6.5cm},clip,width=.85\textwidth]
			{datasets_encoders_across_dims_XGB_clean}
		\end{minipage}
	\end{figure}

	\subsubsection{Prediction with curated data}
	
	We now test the robustness of the different encoding methods to situations where there
	is no need to capture subword information---\textit{e.g.}, low
	cardinality categorical variables, or variables as "Country name", where the
	overlap of character n-grams does not have a relevant meaning.
	We benchmark in \autoref{fig:encoders_benchmark_curated} all encoders on 7 curated
	datasets.
	To simulate black-box usage,
	the dimensionality was fixed to $d{=}30$ for all of them, with the exception of one-hot.
	None of the n-gram based encoders perform worst than one-hot. Indeed,
	the F statistics for the average ranking does not reject the null hypothesis of
	all encoders coming from the same distribution (p-value equal to 0.37).
	
	\subsubsection{Interpretable data science with the Gamma-Poisson}
	
\begin{figure}[t]
		\centering
		\includegraphics[trim={0cm .15cm 0cm 0cm},clip,width=.48\textwidth]
		{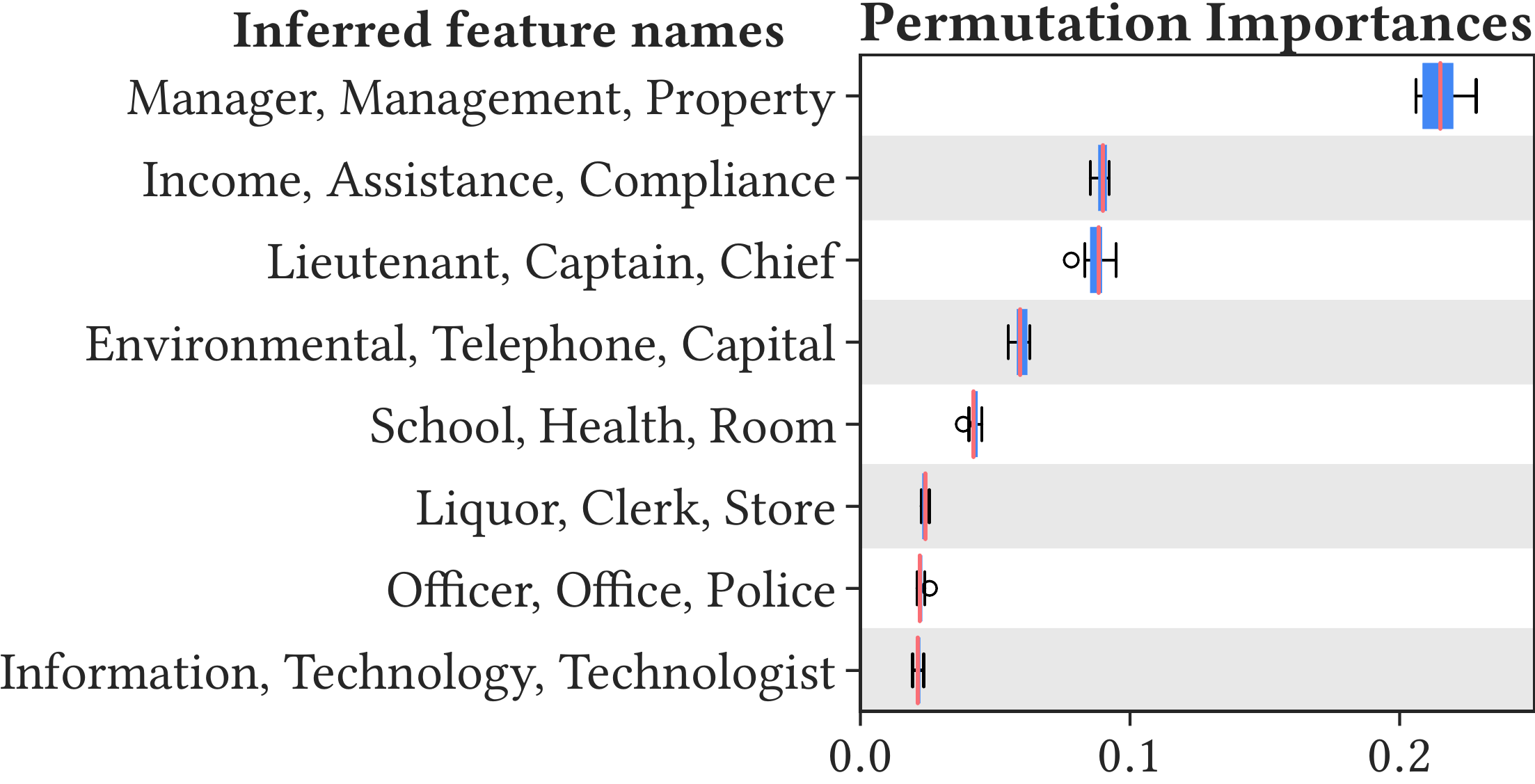}%
		\caption{\textbf{Gamma-Poisson enables interpretable data
			science.}
			The box plots display permutation importances for the variable
			\textit{Employee Position Title} in the Employee Salaries dataset.
			Here we show the 8 most important latent topics from a total of 30.
			The overall feature importances for every feature in the dataset
			are shown in \autoref{fig:permutation_importances_all} in the
			Appendix.}
		\label{fig:permutation_importances_position_title} 
	\end{figure}

	As shown in \autoref{fig:loadings_encoders_simulation}, the Gamma-Poisson
	factorization creates sparse, non-negative feature vectors that are easily
	interpretable as a linear combination of latent categories. We
	give informative features names to each of these latent categories (see
	\ref{sec:inferring_feature_names}). To illustrate how such
	encoding can be used in a data-science setting where humans need to
	understand results,
	\autoref{fig:permutation_importances_position_title} shows the
	permutation
	importances \cite{altmann2010permutation} of each encoding direction of the Gamma-Poisson factorization and its
	corresponding feature names. By far, the most important inferred feature name to predict salaries
	in the Employee Salaries dataset is the latent category
	\texttt{Manager, Management, Property}, which matches general
	intuitions on salaries.

	\section{Discussion and conclusion}
	\label{sec:discussion}
	
	One-hot encoding is not well suited to columns of a table containing
	categories represented with many different strings \cite{cerda2018similarity}.
	Character n-gram count vectors can represent strings well,
	but they dilute the notion of categories with
	extremely high-dimensional vectors.
	A good encoding should capture string similarity between entries and reflect
	it in a lower dimensional encoding.
	
	We study several encoding approaches to capture the structural
	similarities of string entries. 
	The min-hash encoder gives a stateless representation of strings to a
	vector space, transforming inclusions between strings into
	simple inequalities (Theorem \ref{thm:inclusion_rules}).
	A Gamma-Poisson factorization on the count matrix of sub-strings
	gives a low-rank approximation of similarities.
	
	\paragraph*{\bf Scalability}
	Both Gamma-Poisson factorization and the min-hash encoder can be
	used on very large datasets, as they work in streaming
	settings. They markedly improve upon one-hot encoder for large
	scale-learning as \emph{i)} they do not need the definition of a
	vocabulary, \emph{ii)} they give low dimensional representations,
	and thus decrease the cost of subsequent analysis steps.
	Indeed, for both of these encoding approaches, the cost of encoding
	is usually significantly smaller than that of running a powerful
	supervised learning method such as XGBoost, even on the reduced
	dimensionality (\autoref{tab:times_GammaPoisson_XGB30} in 
	the Appendix).
	The min-hash encoder is unique in terms of scalability, as it
	gives low-dimensional representations while being
        completely stateless, which greatly facilitates distributed
	computing. The representations enable much better statistical
        analysis than a simpler stateless low-dimensional encoding built
	with random projections of n-gram string representations.
	Notably, the most scalable encoder is also the best performing
	for supervised learning, at the cost of some loss in
	interpretability.
	
\begin{figure*}[p]
		\begin{subfigure}[t]{1\textwidth}
			\raggedleft
			\includegraphics[trim={0cm .3cm 0cm 0cm},clip,width=1\textwidth]
			{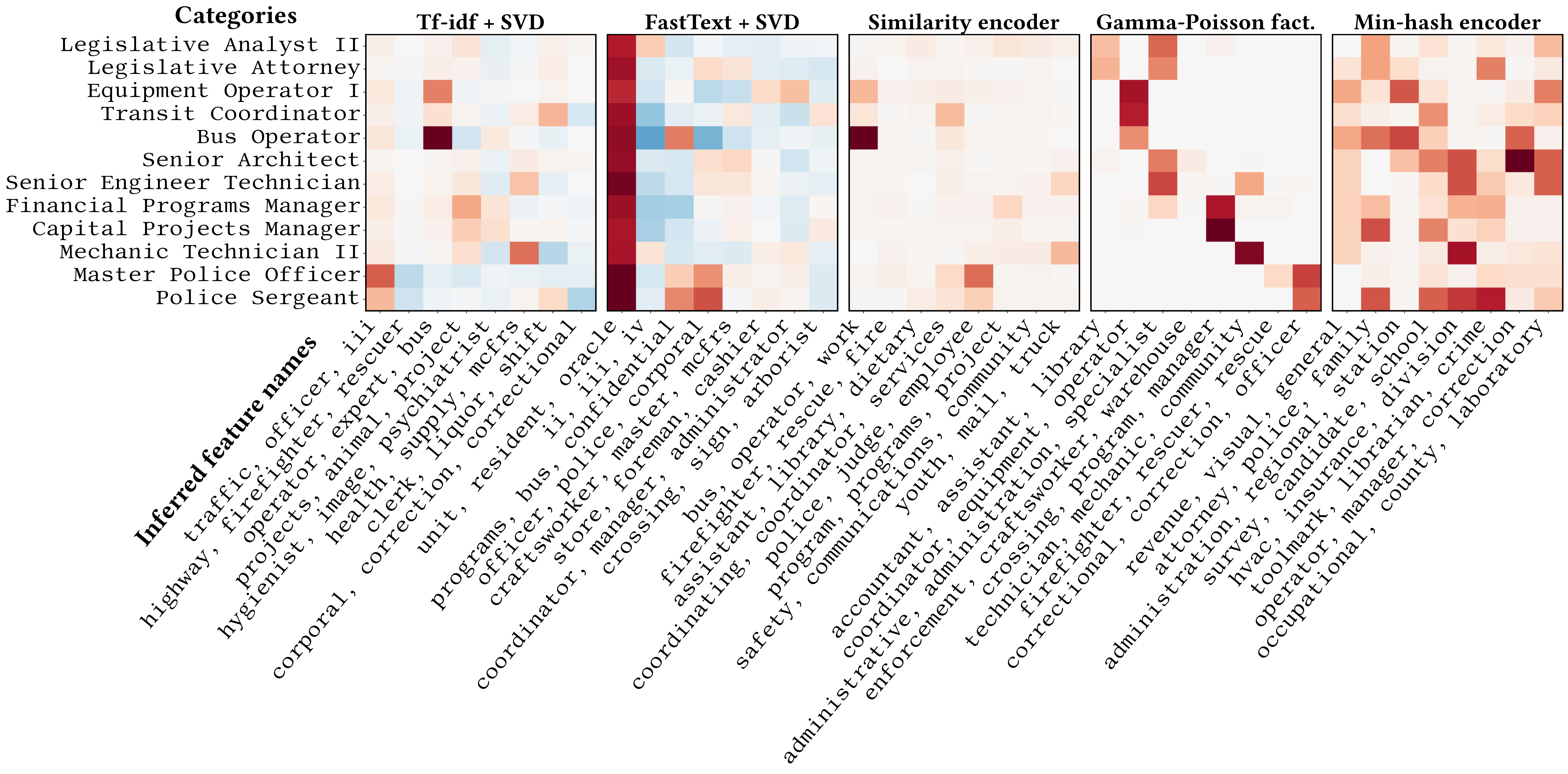}
			\caption{\textit{Employee Position Title} (Employee Salaries dataset)}
		\end{subfigure}
		
		\vspace{.3cm}
		\begin{subfigure}[t]{1\textwidth}
			\raggedleft
			\includegraphics[trim={0 .3cm 0 0cm},clip,height=.122\textheight]
			{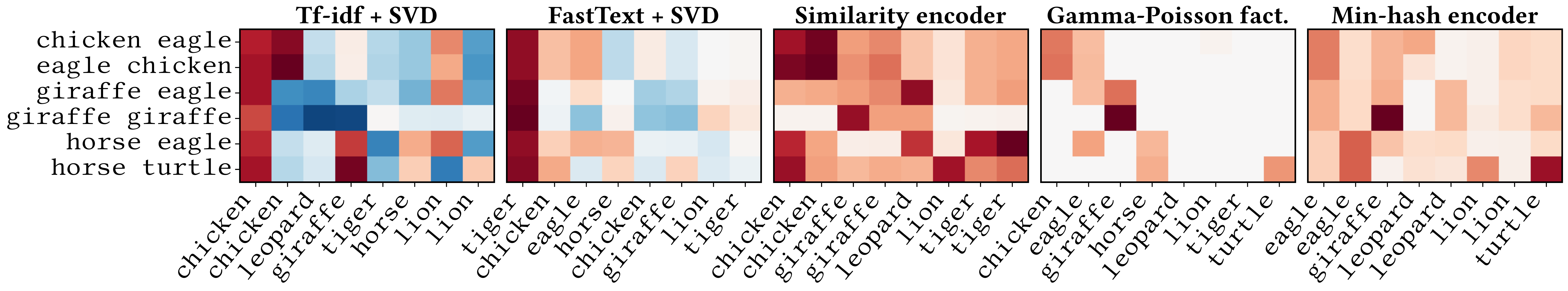}
			\caption{Simulated multi-label entries}
		\end{subfigure}%
		
		\vspace{.3cm}
		\begin{subfigure}[t]{\textwidth}
			\includegraphics[trim={0 0cm 66.2cm 12cm},clip,height=.08\textheight]
			{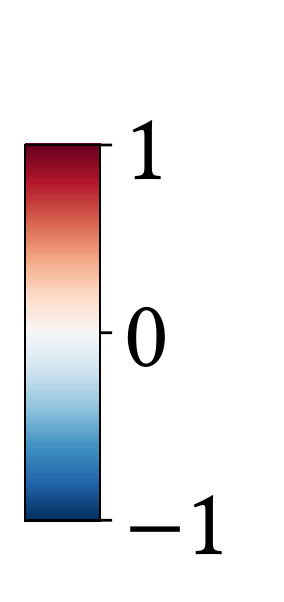}%
			\hfill
			\includegraphics[trim={1cm 5.55cm 0 0cm},clip,height=.122\textheight]
			{loadings_simulation_typos_appendix}
			\caption{Simulated entries with typos}
		\end{subfigure}%
		
		\caption{\textbf{The Gamma-Poisson factorization gives positive and
				sparse representations that are easily interpretable.}
			Encoding vectors (d=8) for simulated (a and b) and a real dataset (c)
			obtained with different encoding methods for some categories
			($y$-axis).
			The $x$-axis shows the activations with their respective inferred
			feature names.}
		\label{fig:loadings_encoders_simulation_appendix} 
	\end{figure*}

	\paragraph*{\bf Recovery of latent categories}
	Describing results in terms of a small number of categories can
	greatly help interpreting a statistical analysis. Our experiments
	on real and simulated data show that encodings created by the
	Gamma-Poisson factorization correspond to loadings on meaningful recovered
	categories. It removes the need to manually curate entries to
	understand what drives an analysis. For this, positivity of the
	loadings and the soft sparsity imposed by the Gamma prior is
	crucial; a simple SVD fails to give interpretable loadings (Appendix
	\autoref{fig:loadings_encoders_simulation_appendix}).
		
	\paragraph*{\bf AutoML settings}
	AutoML (automatic machine learning) strives to develop
	machine-learning pipeline that can be applied to datasets without
	human intervention
	\cite{hutter2015automatic,hutter2019automated}. To date, it has
	focused on tuning and model selection for supervised learning on
	numerical data. Our work addresses the feature-engineering step.
	In our experiments, we apply the exact same prediction pipeline
	to 17 non-curated and 7 curated tabular datasets, without any
	custom feature engineering. Both Gamma-Poisson factorization and
	min-hash encoder led to best-performing prediction accuracy,
	using a classic gradient-boosted tree implementation (XGBoost).
	We did not tune hyper-parameters of the encoding, such as
	dimensionality or parameters of the priors for the Gamma Poisson.
	They adapt to the language and the vocabulary of the entries,
	unlike NLP embeddings such as fastText which must have been
previously extracted on a corpus of the language (\autoref{fig:fasttext_language_comparison}).
	These string categorical encodings therefore open the door to
	autoML on the original data, removing the need for feature
	engineering which can lead to difficult model selection.
	A
	possible rule when
	integrating tabular data into an autoML pipeline could be to
	apply min-hash or Gamma-Poisson encoder for string categorical
	columns with a cardinality above 30, and use one-hot encoding for
	low-cardinality columns. Indeed, results show that these
	encoders are also suitable for normalized entries.\\
	
	One-hot encoding is the defacto standard for statistical analysis
	on categorical entries. Beyond its simplicity, its strength is to
	represent the discrete nature of categories. However, it becomes
	impractical when there are too many different unique entries, for
	instance because the string representations have not been curated
	and display typos or combinations of multiple informations in the
	same entries. For high-cardinality
	string categories, we have presented two scalable approaches to
	create low-dimensional encoding that retain the qualitative
	properties of categorical entries. The min-hash encoder is
	extremely scalable and gives the best prediction performance
	because it transforms string inclusions to
	vector-space operations that can easily be captured by a
	supervised learning step.
	If interpretability of results is an issue,
	the Gamma-Poisson factorization performs almost as well for
	supervised learning, but enables expressing results in terms of
	meaningful latent categories.
	As such, it gives a readily-usable
	replacement to one-hot encoding for high-cardinality string categorical
	variables. Progress brought by these encoders is important, as
	they avoid one of the time-consuming steps of data science:
	normalizing entries of databases via human-crafted rules.
	
	\ifCLASSOPTIONcompsoc
	\section*{Acknowledgments}
	\else
	\section*{Acknowledgment}
	\fi
	Authors were supported by the DirtyData (ANR-17-CE23-0018-01) and
	the FUI Wendelin projects.
	
	\ifCLASSOPTIONcaptionsoff
	\newpage
	\fi
	
	\bibliographystyle{IEEEtran}
	\bibliography{biblio}
	%
	
	%
	
	\begin{IEEEbiography}[{\includegraphics[width=1in,height=1.25in,trim={.7cm
				0cm .7cm 0cm},clip,keepaspectratio]{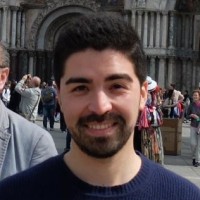}}]{Patricio Cerda}
		Patricio holds a masters degree in applied mathematics
		from École Normale Supérieure Paris-Saclay and a PhD in
		computer science from Université Paris-Saclay.
		His research interests are natural language processing,
		econometrics and causality.s
		
	\end{IEEEbiography}
	\begin{IEEEbiography}[{\includegraphics[width=1in,height=1.25in,trim={5cm
0 5cm 0},clip,keepaspectratio]{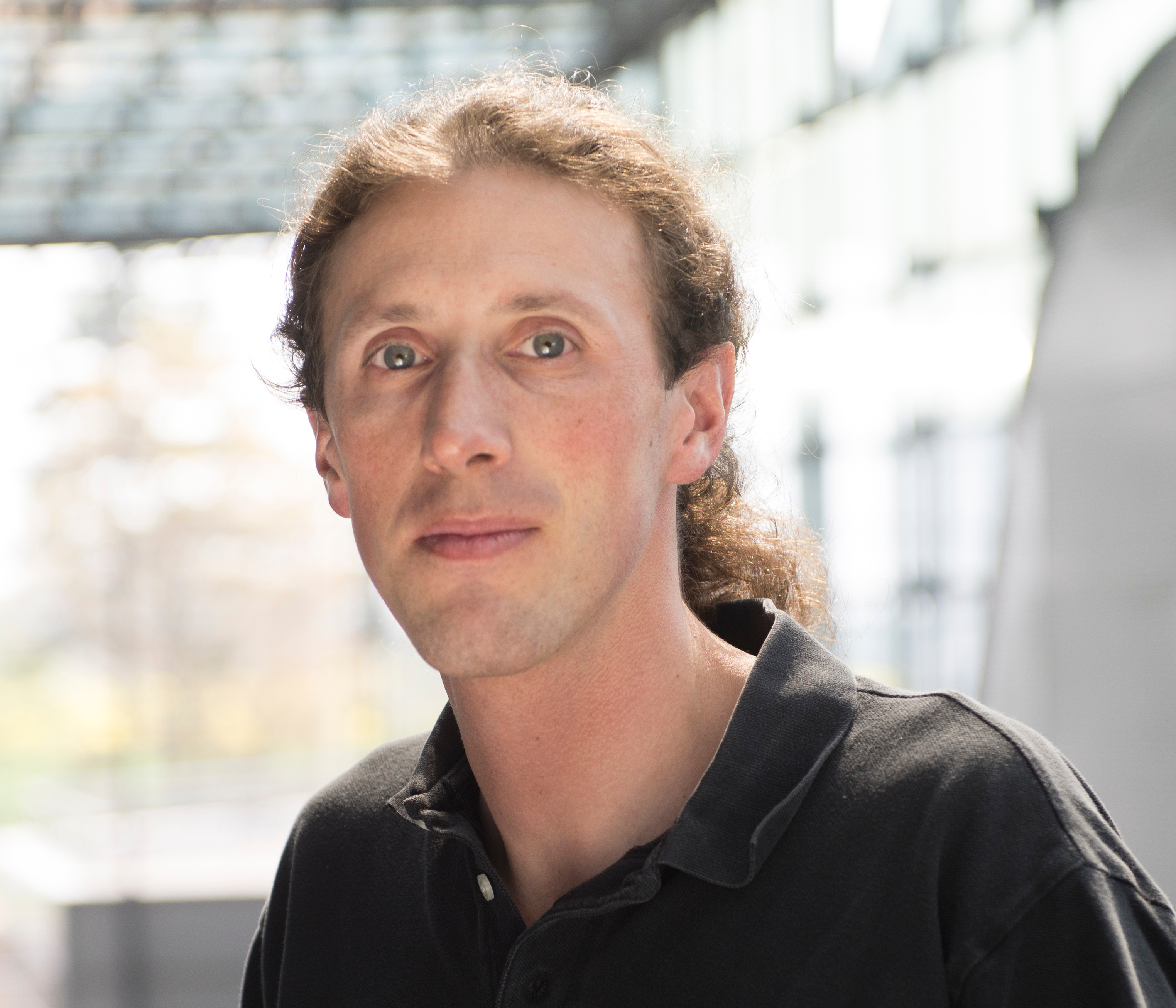}}]{Gaël Varoquaux}
		Gaël Varoquaux is a research director at Inria
		developing statistical learning for data science
		and scientific inference.
		He has pioneered machine learning on
		brain images.
		More generally, he develops tools to make machine learning easier,
			for real-life, uncurated data.
		He co-funded scikit-learn and helped build central tools for data analysis
		in Python.
		He has a PhD in quantum physics and graduated from
		École Normale Supérieure Paris.
	\end{IEEEbiography}
	\vfill
	
	
	
	
	

	
	%
	
	\newpage
	\appendices
	\section{Reproducibility}
	\subsection{Dataset Description}
	
	\subsubsection{Non-curated datasets}
	\label{sec:dataset_description}
	
	\hspace{.4cm}
	\textbf{Building Permits}\footnote{
		\url{https://www.kaggle.com/chicago/chicago-building-permits}}
	(sample size: 554k).
	Permits issued by the Chicago Department of Buildings since 2006.
	Target (regression): \emph{Estimated Cost}.
	Categorical variable: \emph{Work Description} (cardinality: 430k).
	
	\textbf{Colleges}\footnote{
		\url{https://beachpartyserver.azurewebsites.net/VueBigData/DataFiles/Colleges.txt}} (7.8k).
	Information about U.S. colleges and schools.
	Target (regression): \emph{Percent Pell Grant}.
	Cat. var.: \emph{School Name} (6.9k).
	
	\textbf{Crime Data}\footnote{
		\url{https://data.lacity.org/A-Safe-City/Crime-Data-from-2010-to-Present/y8tr-7khq}} (1.5M).
	Incidents of crime in the City of Los Angeles since 2010.
	Target (regression): \emph{Victim Age}.
	Categorical variable: \emph{Crime Code Description} (135).
	
	\textbf{Drug Directory}\footnote{
		\url{https://www.fda.gov/Drugs/InformationOnDrugs/ucm142438.htm}}
	(120k).
	Product listing data submitted to the U.S. FDA for all unfinished,
	unapproved drugs.
	Target (multiclass): \emph{Product Type Name}.
	Categorical var.: \emph{Non Proprietary Name} (17k).
	
	\textbf{Employee Salaries}\footnote{
		\url{https://catalog.data.gov/dataset/employee-salaries-2016}} (9.2k).
	Salary information for employees of the
	Montgomery County, MD.
	Target (regression): \emph{Current Annual Salary}.
	Categorical variable: \emph{Employee Position Title} (385).
	
	\textbf{Federal Election}\footnote{
		\url{https://classic.fec.gov/finance/disclosure/ftpdet.shtml}}
	(3.3M).
	Campaign finance data for the 2011-2012 US election cycle.
	Target (regression): \emph{Transaction Amount}.
	Categorical variable: \emph{Memo Text} (17k).
	
	\textbf{Journal Influence}\footnote{
		\url{https://github.com/FlourishOA/Data}} (3.6k).
	Scientific journals and the respective influence scores.
	Target (regression): \emph{Average Cites per Paper}.
	Categorical variable: \emph{Journal Name} (3.1k).
	
	\textbf{Kickstarter Projects}\footnote{
		\url{https://www.kaggle.com/kemical/kickstarter-projects}} (281k).
	More than 300,000 projects from
	\url{https://www.kickstarter.com}.
	Target (binary): \emph{State}.
	Categorical variable: \emph{Category} (158).
	
	\textbf{Medical Charges}\footnote{
		\url{https://www.cms.gov/Research-Statistics-Data-and-Systems/Statistics-Trends-and-Reports/Medicare-Provider-Charge-Data/Inpatient.html}}
	(163k).
	Inpatient discharges for Medicare
	beneficiaries for more than 3,000 U.S. hospitals.
	Target (regression): \emph{Average Total Payments}.
	Categorical var.:
	\emph{Medical Procedure} (100). 
	
	\textbf{Met Objects}\footnote{
		\url{https://github.com/metmuseum/openaccess}} (469k).
	Information on artworks objects of the Metropolitan Museum of Art's collection.
	Target (binary): \emph{Department}.
	Categorical variable: \emph{Object Name} (26k).
	
	\textbf{Midwest Survey}\footnote{
		\url{https://github.com/fivethirtyeight/data/tree/master/region-survey}}
	(2.8k).
	Survey
	to know if people self-identify as Midwesterners.
	Target (multiclass):
	\emph{Census Region} (10 classes).
	Categorical var.: \emph{What would you call the part
		of the country you live in now?} (844).
	
	\textbf{Open Payments}\footnote{
		\url{https://openpaymentsdata.cms.gov}}
	(2M).
	Payments given by healthcare
	manufacturing companies to medical doctors or hospitals (year 2013).
	Target (binary): \emph{Status} (if the payment was made
	under a research protocol).
	Categorical var.: \emph{Company name} (1.4k).
	
	\textbf{Public Procurement}\footnote{
		\url{https://data.europa.eu/euodp/en/data/dataset/ted-csv}}
	(352k).
	Public procurement data for the European Economic Area, Switzerland,
	and the Macedonia.
	Target (regression): \emph{Award Value Euro}.
	Categorical var.: \emph{CAE Name} (29k).
	
	\textbf{Road Safety}\footnote{
		\url{https://data.gov.uk/dataset/road-accidents-safety-data}}
	(139k).
	Circumstances of personal injury of road accidents in Great Britain from 1979. Target (binary): \emph{Sex of Driver}.
	Categorical variable: \emph{Car Model} (16k).
	
	\textbf{Traffic Violations}\footnote{
		\url{https://catalog.data.gov/dataset/traffic-violations-56dda}}
	(1.2M).
	Traffic information from electronic
	violations issued in the  Montgomery County, MD.
	Target (multiclass): \emph{Violation type} (4 classes).
	Categorical var.: \emph{Description} (11k).
	
	\textbf{Vancouver Employee}\footnote{
		\url{https://data.vancouver.ca/datacatalogue/employeeRemunerationExpensesOver75k.htm}}(2.6k).
	Remuneration and expenses for employees earning over \$75,000 per year.
	Target (regression): \emph{Remuneration}.
	Categorical variable: \emph{Title} (640).
	
	\textbf{Wine Reviews}\footnote{
		\url{https://www.kaggle.com/zynicide/wine-reviews/home}} (138k).
	Wine reviews scrapped from WineEnthusiast.
	Target (regression): \emph{Points}.
	Categorical variable: \emph{Description} (89k).
	
	\subsubsection{Curated datasets}
	\label{subsubsec:description_curated_datasets}
	\hspace{.4cm}
	\textbf{Adult}\footnote{
		\url{https://archive.ics.uci.edu/ml/datasets/adult}} (sample size: 32k).
	Predict whether income exceeds \$50K/yr based on census data.
	Target (binary): \emph{Income}.
	Categorical variable: \emph{Occupation} (cardinality: 15).
	
	\textbf{Cacao Flavors}\footnote{
		\url{https://www.kaggle.com/rtatman/chocolate-bar-ratings}} (1.7k).
	Expert ratings of over 1,700 individual chocolate bars, along with information on their origin and bean variety.
	Target (multiclass): \emph{Bean Type}.
	Categorical variable: \emph{Broad Bean Origin} (100).
	
	\textbf{California Housing}\footnote{
		\url{https://github.com/ageron/handson-ml/tree/master/datasets/housing}} (20k).
	Based on the 1990 California census data. It contains one row per census block group (a block group typically has a population of 600 to 3,000 people).
	Target (regression): \emph{Median House Value}.
	Categorical variable: \emph{Ocean Proximity} (5).
	
	\textbf{Dating Profiles}\footnote{
		\url{https://github.com/rudeboybert/JSE_OkCupid}} (60k).
	Anonymized data of dating profiles from OkCupid.
	Target (regression): \emph{Age}.
	Categorical variable: \emph{Diet} (19).
	
	\textbf{House Prices}\footnote{
		\url{https://www.kaggle.com/c/house-prices-advanced-regression-techniques}} (1.1k).
	Contains variables describing residential homes in Ames, Iowa.
	Target (regression): \emph{Sale Price}.
	Categorical variable: \emph{MSSubClass} (15).
	
	\textbf{House Sales}\footnote{
		\url{https://www.kaggle.com/harlfoxem/housesalesprediction}} (21k).
	Sale prices for houses in King County, which includes Seattle.
	Target (regression): \emph{Price}.
	Categorical variable: \emph{ZIP code} (70).
	
	\textbf{Intrusion Detection}\footnote{
		\url{https://archive.ics.uci.edu/ml/datasets/KDD+Cup+1999+Data}} (493k).
	Network intrusion simulations with a variaty od descriptors of the attack type.
	Target (multiclass): \emph{Attack Type}.
	Categorical variable: \emph{Service} (66).
	
	\subsection{Learning pipeline}
	\label{sec:pipeline}
	
	\paragraph*{\textbf{Sample size}} Datasets' size range from a couple of
	thousand to several million samples.
	To reduce computation time on the learning step,
	the number of samples was limited to 100k for large datasets.
	
	\paragraph*{\textbf{Data preprocessing}} We removed rows with missing values in the
	target or in any explanatory variable other than the selected
	categorical variable, for which we replaced missing entries by the string `nan'.
	The only additional preprocessing for the categorical variable was to transform
	all entries to lower case.
	
	\paragraph*{\textbf{Cross-validation}} For every dataset, we made 20 random
	splits of the data, with one third of samples for testing at each time. In the
	case of binary classification, we performed stratified randomization.
	
	\paragraph*{\textbf{Performance metrics}}
	Depending on the type of prediction task,
	we used different scores to evaluate the performance of the supervised
	learning problem:
	for regression, we used the $R^2$ score; for binary classification,
	the average precision; and for multi-class classification, the accuracy score.
	
	\paragraph*{\textbf{Parametrization of classifiers}}
	We used the scikit-learn \cite{pedregosa2011scikit} for most
	of the data processing. For all the experiments, we used the scikit-learn
	compatible implementations of XGBoost \cite{chen2016xgboost}, with a grid search
	on the {\small\verb|learning_rate|} (0.05, 0.1, 0.3) and
	{\small\verb|max_depth|} (3, 6, 9) parameters.
	All datasets and encoders use the same parametrization.
	
	\paragraph*{\textbf{Dimensionality reduction}}
	We used the scikit-learn
	implementations of {\small \verb|TruncatedSVD|}
	and {\small \verb|GaussianRandomProjection|}, with the default
	parametrization in both cases.
	
	\subsection{Synthetic data generation}
	\label{sec:synthetic_data}
	
	\paragraph*{\textbf{Multi-label categories}}
	The multi-label data was created by
	concatenating $k{+}2$ ground truth categories (labels), with $k$
	following a Poisson distribution---hence,
	all entries contain at least two concatenated labels.
	Not having single labels in the synthetic data
	makes the recovering of latent categories harder. 
	
	\paragraph*{\textbf{Typo generator}}
	For the simulation of typos, we added 10\% of variations of the
	original ground truth categories by adding errors randomly
	(missing, swaped, inserted and replaced characters). For each
	ground-truth category, a list of misspelled candidates (at least 15 per
	category) was obtained from the website:
	\url{https://www.dcode.fr/typing-error-generator}.
	Then, the misspelled categories were randomly chosen
	to generate the 10\% of typos.

	\subsection{Online Resources}
	\label{sec:online_resources}
	
	Experiments are available in Python code at
	\url{https://github.com/pcerda/string_categorical_encoders}.
	Implementations and examples on learning with string categories
	can be found at \url{http://dirty-cat.github.io}.
	The available encoders are compatible with the scikit-learn's API.

	\section{Algorithmic considerations}
	
	\begin{figure}[t!]
		\centering
		\includegraphics[trim={0 0 0 0},clip,height=.21\textheight]
		{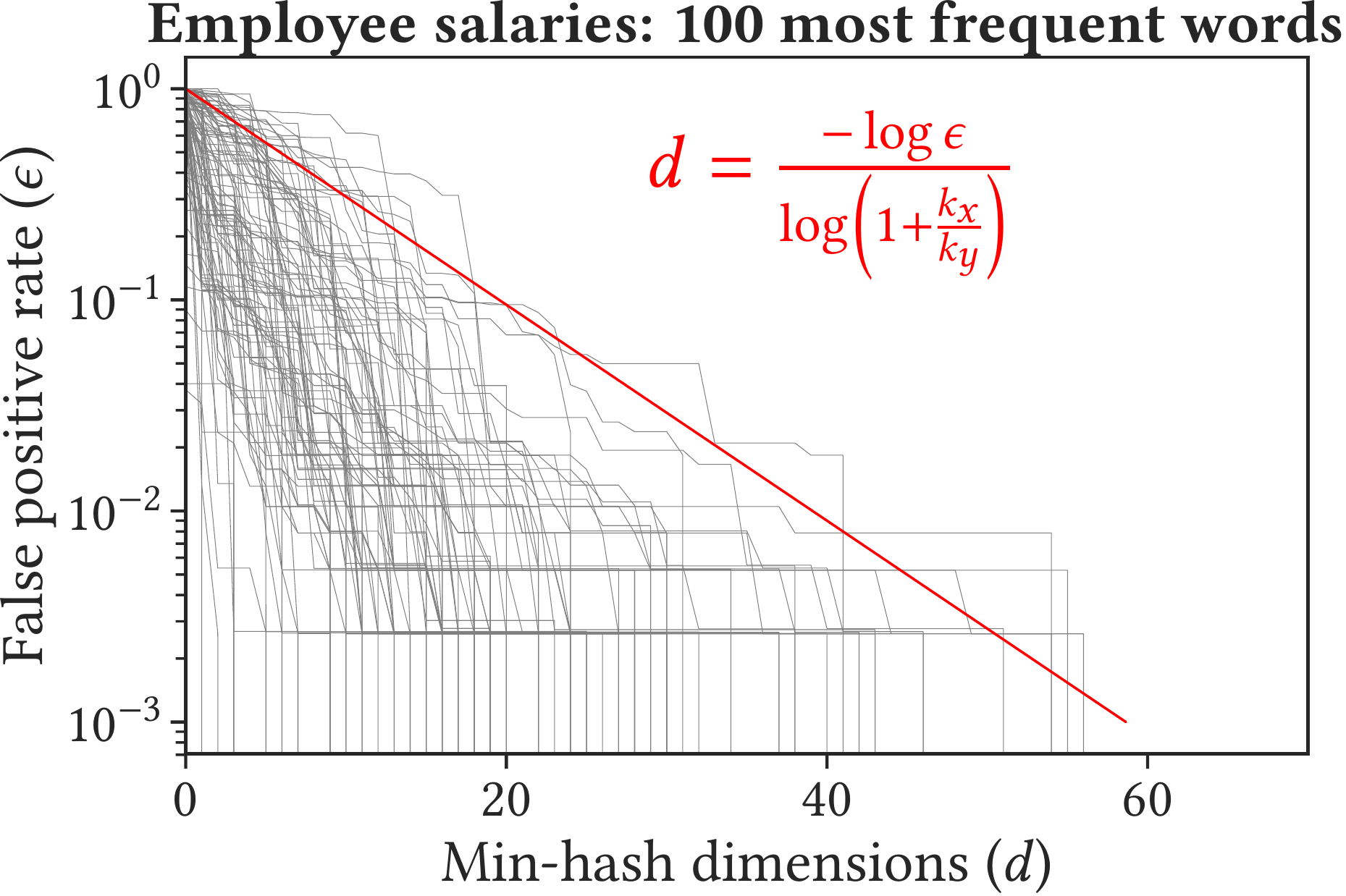}
		\caption{\textbf{Number of dimensions required to identify inclusions.}
			Grey lines are the proportion of false positives obtained
			for the 100 most frequent words in the employee salaries dataset
			($H_0$ corresponds to identifying categories that do not contain the
			given word).
			The red line represents the theoretical minimum dimensionality
			required to obtain a desired false positive rate
			(with $k_x {/} k_y = 0.125$, the inverse of the maximum number of
			words per category), as shown in Theorem \ref{thm:inclusion_rules}.}
		\label{fig:employee_salaries_false_positives}
	\end{figure}
	
	\subsection{Gamma-Poisson factorization}
	
	\begin{table}[h]
		\caption{Parameter values for the Gamma-Poisson factorization.
		The same parameters were used for all datasets.}
		\label{tab:parameters_gamma_poisson}       
		\centering
		\normalsize
		\setlength\tabcolsep{6pt}
		\rowcolors{2}{black!5}{white}
		\begin{tabular}{clc}
			\hline\noalign{\smallskip}
			\textbf{Parameter} & \textbf{Definition} & \textbf{Default value} \\
			\hline\noalign{\smallskip}
			$\alpha_i$ & Poisson shape & 1.1	\\
			$\beta_i$ & Poisson scale & 1.0 \\
			$\rho$ & Discount factor &	0.95 \\
			$q$	& Mini-batch size & 256	\\
			$\eta$ & Approximation error & $10^{-4}$ \\
			$\epsilon$ & Approximation error & $10^{-3}$ \\
			\hline
		\end{tabular}
	\end{table}
	
\begin{figure}[p]
		\centering
		\includegraphics[width=.32\textwidth]
		{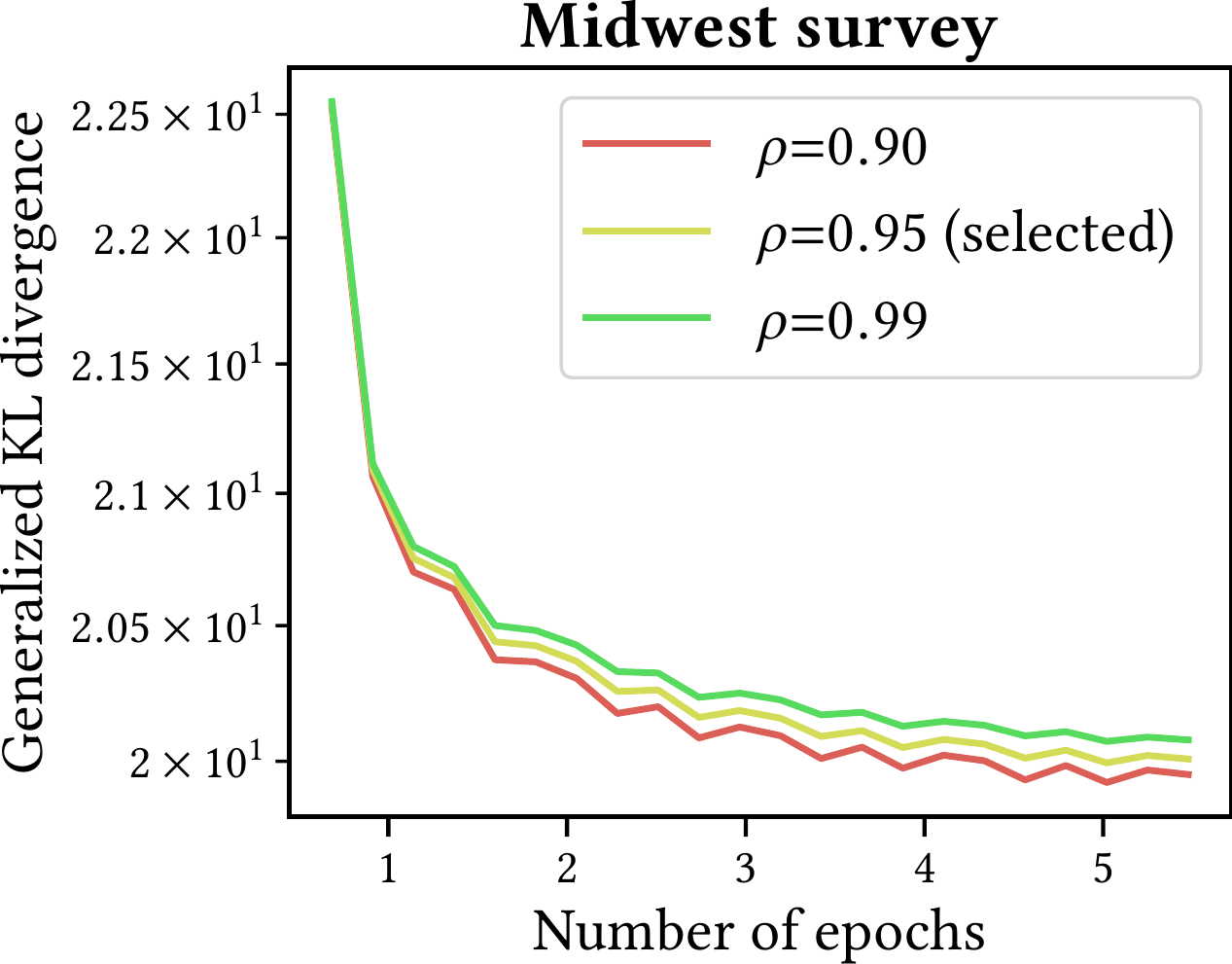}%
		
		\vspace{.3cm}
		\includegraphics[width=.32\textwidth]
		{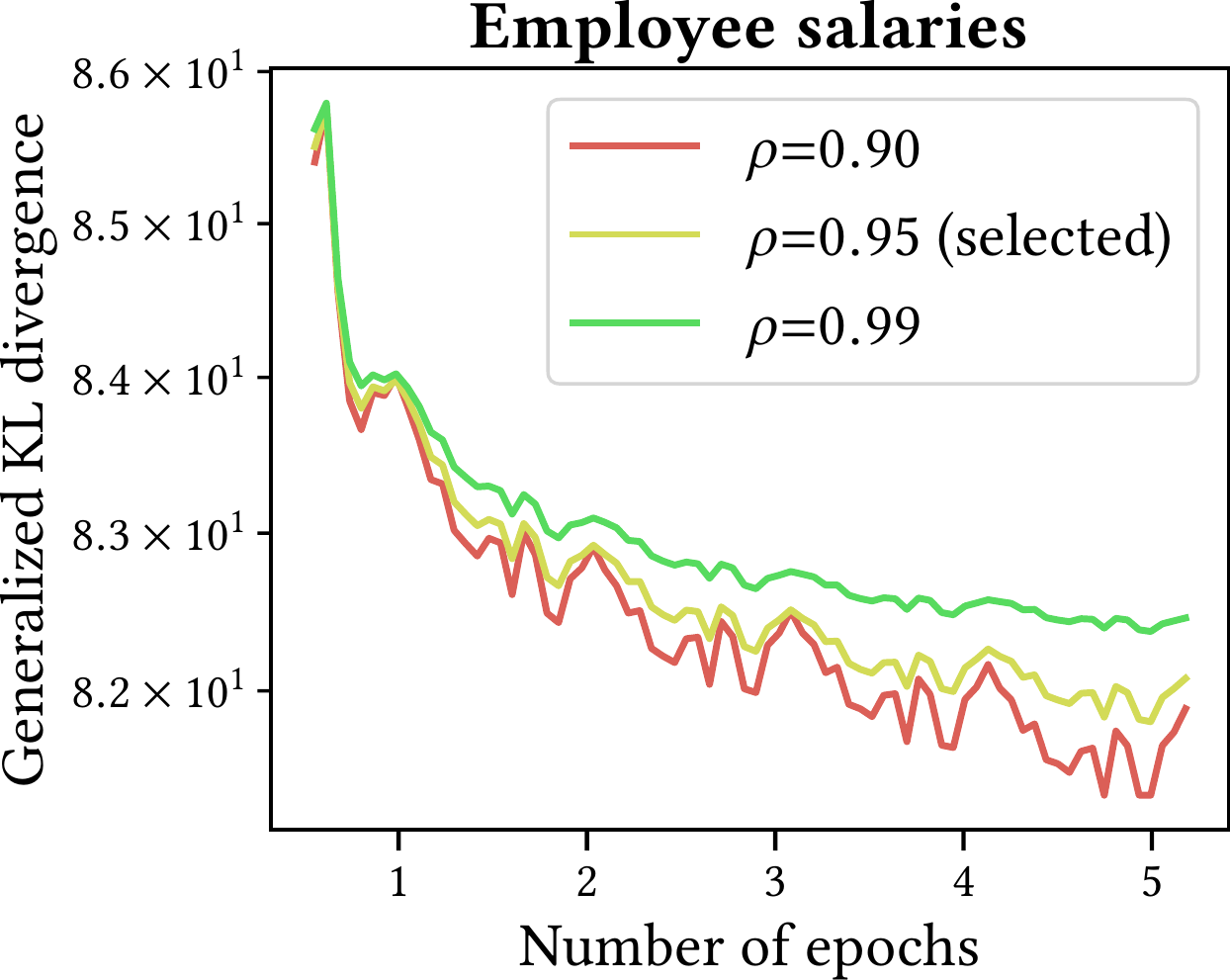}%
		
		\vspace{.3cm}
		\includegraphics[width=.32\textwidth]
		{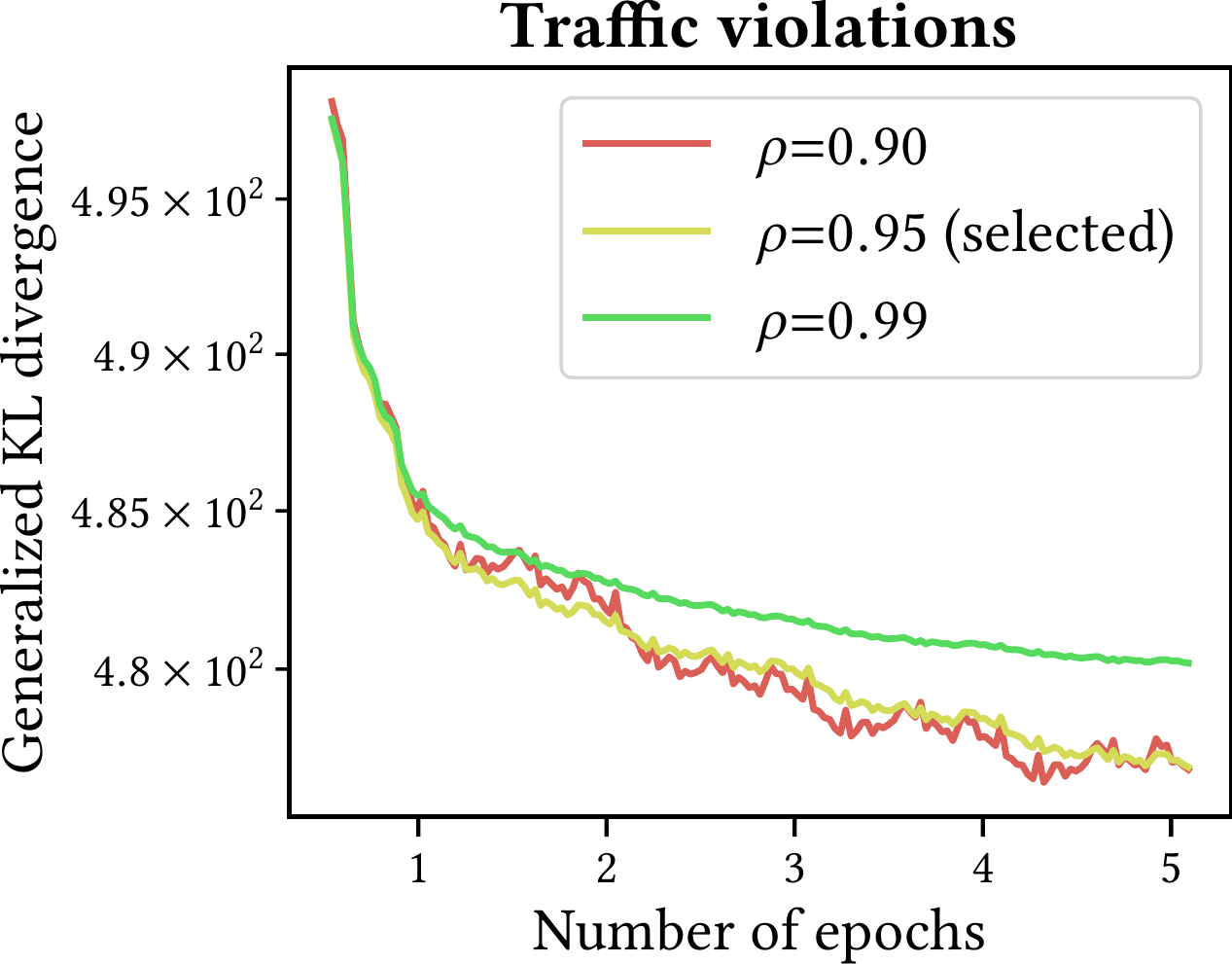}%
		\caption{\textbf{Convergence for different discount factor values for the
			Gamma-Poisson model.}
		In all experiments, the value $\rho = 0.95$ is used, as it gives a good trade-off between convergence and stability of the solution across the number of epochs.
		}
		\label{fig:benchmark_rho_gamma-poisson}       
	\end{figure}

	Algorithm \ref{alg:online_GP} requires some input parameters and initializations that can affect convergence.
	One important parameter is $\rho$, the discount factor for the fitting
	in the past. \autoref{fig:benchmark_rho_gamma-poisson} shows that choosing
	$\rho{=}.95$ gives the best compromise between stability of the convergence
	and data fitting in terms of the Generalized KL divergence.
	The default values used in the experiments are listed in
	\autoref{tab:parameters_gamma_poisson}.
	
\begin{figure}[p]
		\centering
		\includegraphics[width=.32\textwidth]
		{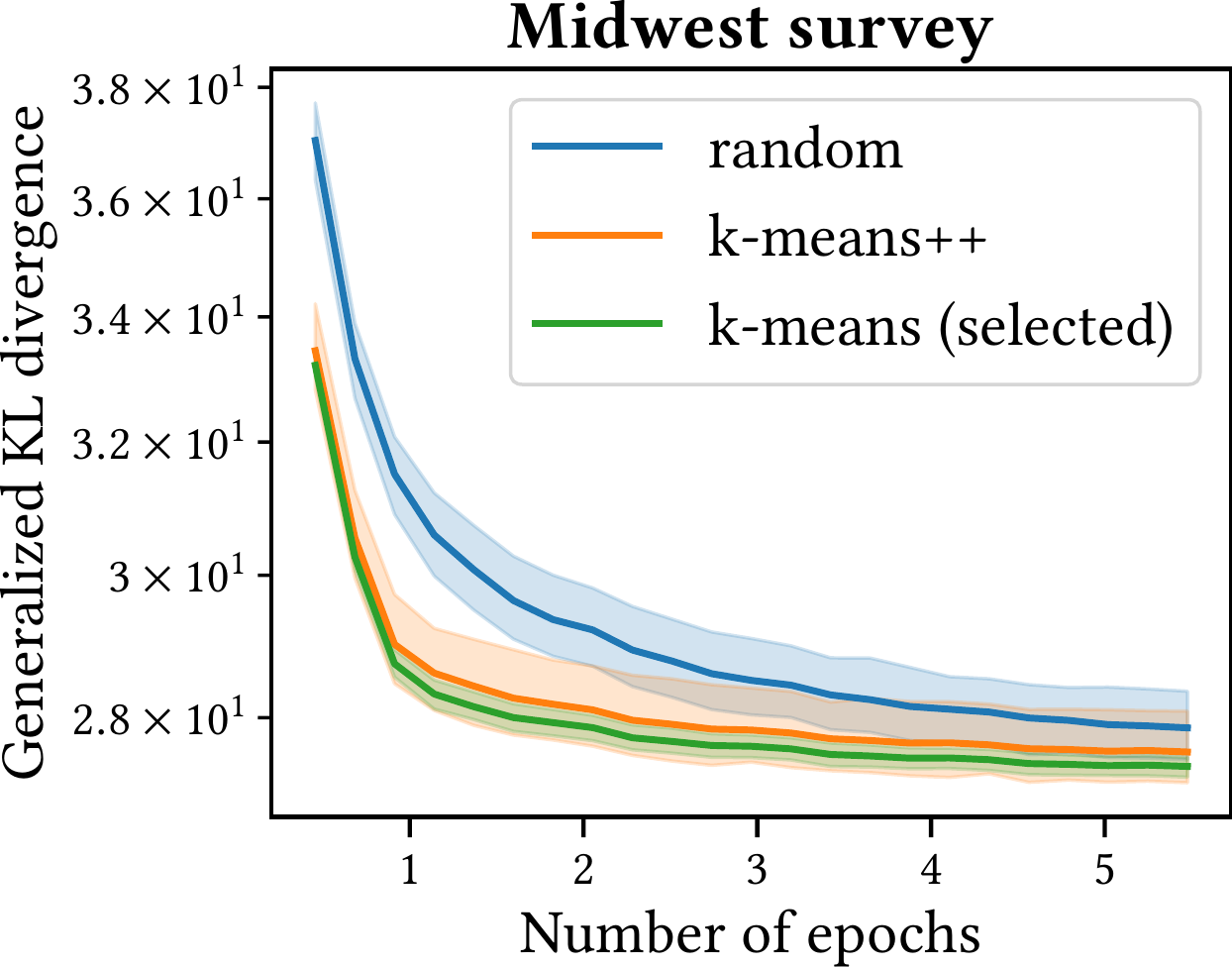}%
		
		\vspace{.3cm}
		\includegraphics[width=.32\textwidth]
		{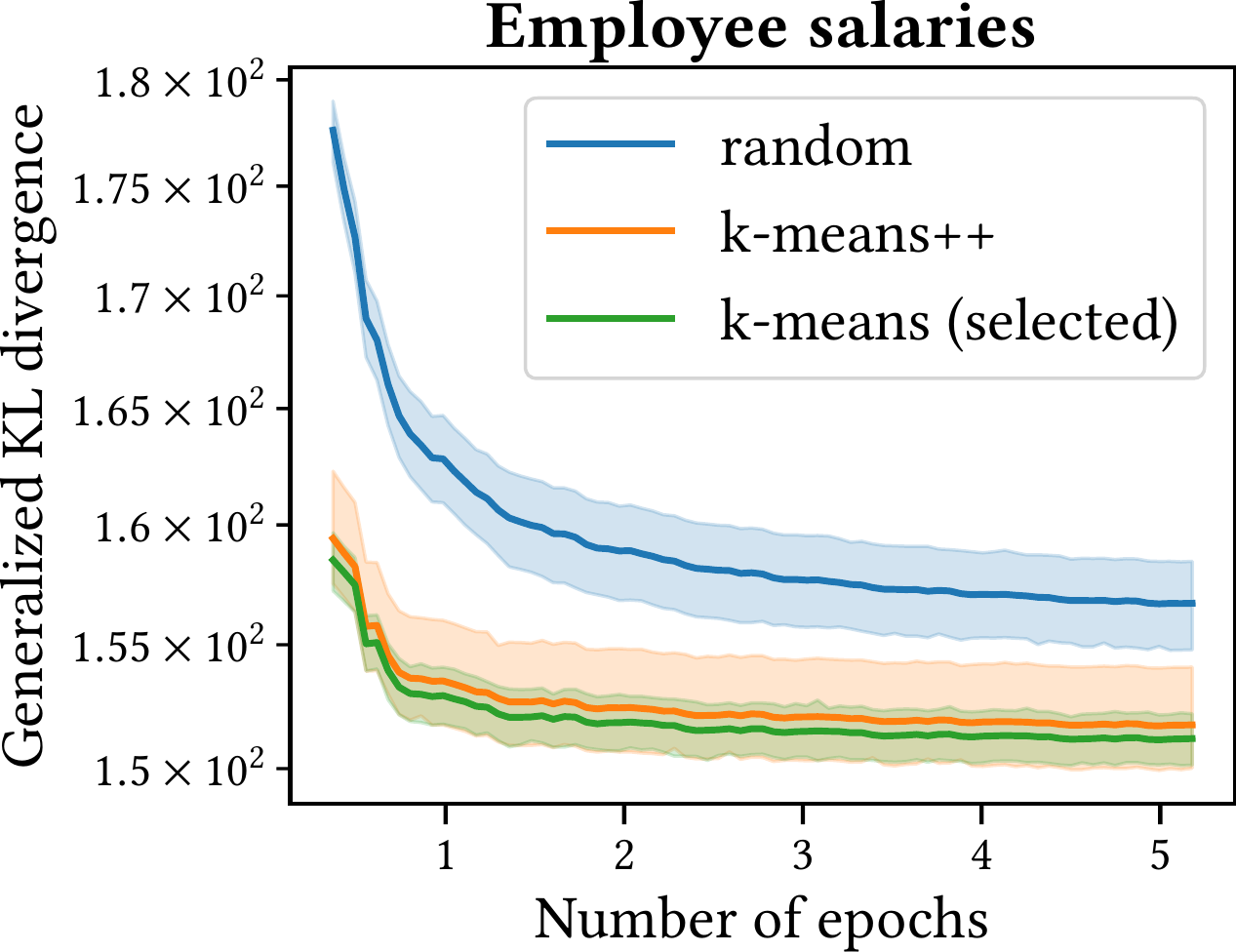}%
		
		\vspace{.3cm}
		\includegraphics[width=.32\textwidth]
		{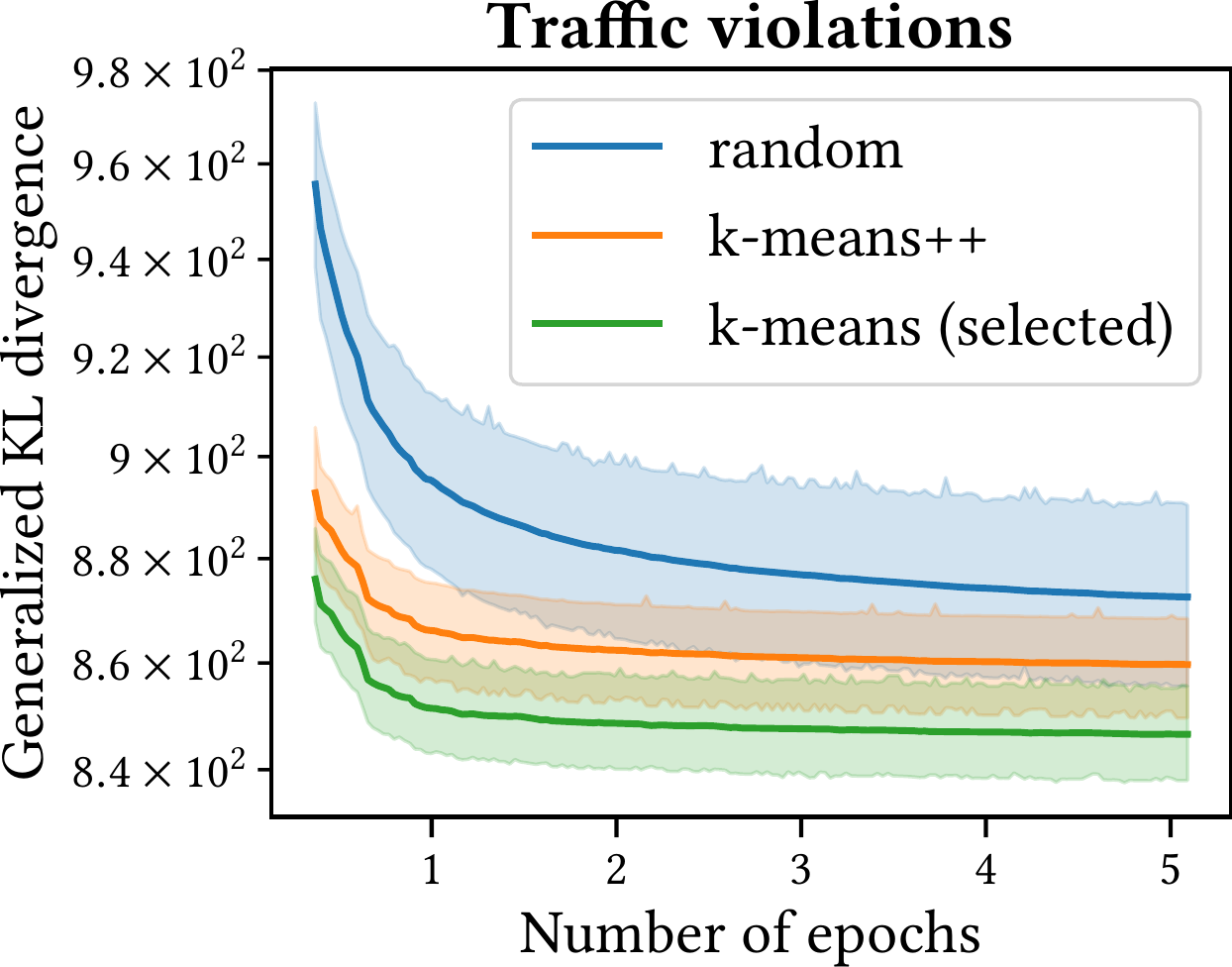}%
		\caption{\textbf{Convergence for different initializations for the Gamma-Poisson model.}
		In all experiments, the k-means strategy is used.}
		\label{fig:benchmark_init_gamma-poisson}       
	\end{figure}

	With respect to the initialization of the topic matrix
	$\mathbf{\Lambda}^{(0)}$,
	the best option is to choose the centroids of a k-means clustering
	(\autoref{fig:benchmark_init_gamma-poisson}) in a
	hashed version of the n-gram count matrix $\mathbf{F}$ in a reduced
	dimensionality (in order to speed-up convergence of the k-means algorithm)
	and then project back to the n-gram space with a nearest neighbors algorithm.

	\section{Additional figures and tables}\label{sec:additional_figures}

\begin{figure}[t]
	\centering
	\includegraphics[trim={0cm 0cm 0cm 0cm},clip,width=.9\linewidth]
	{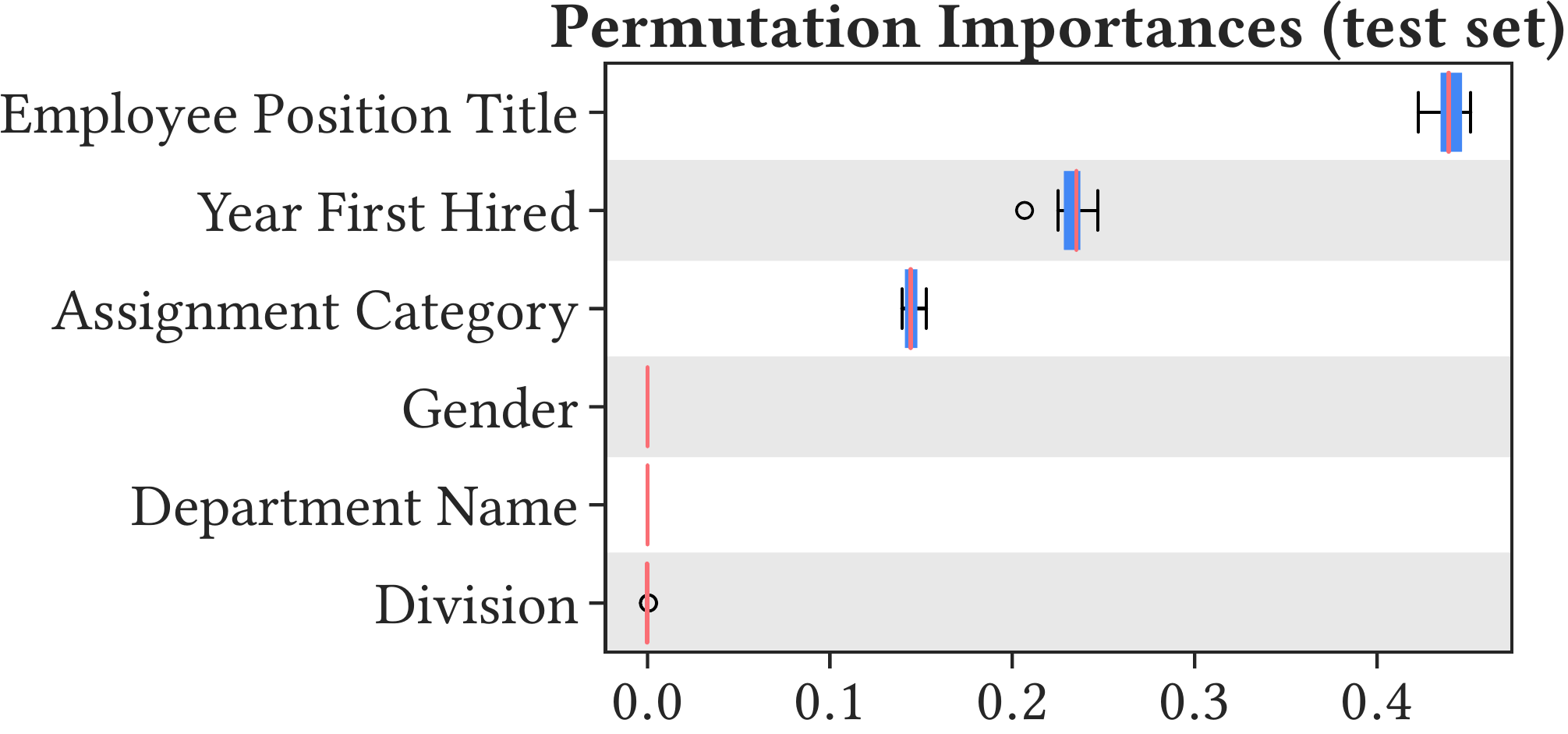}%
	\caption{\textbf{Overall permutation importances for every feature
			in the Employee Salaries dataset.}
	}
	\label{fig:permutation_importances_all} 
\end{figure}

\begin{figure}[p]
	\begin{subfigure}[t]{.33\textwidth}
		\includegraphics[trim={-4cm 0cm 0cm 0cm},clip,height=.22\textheight]
		{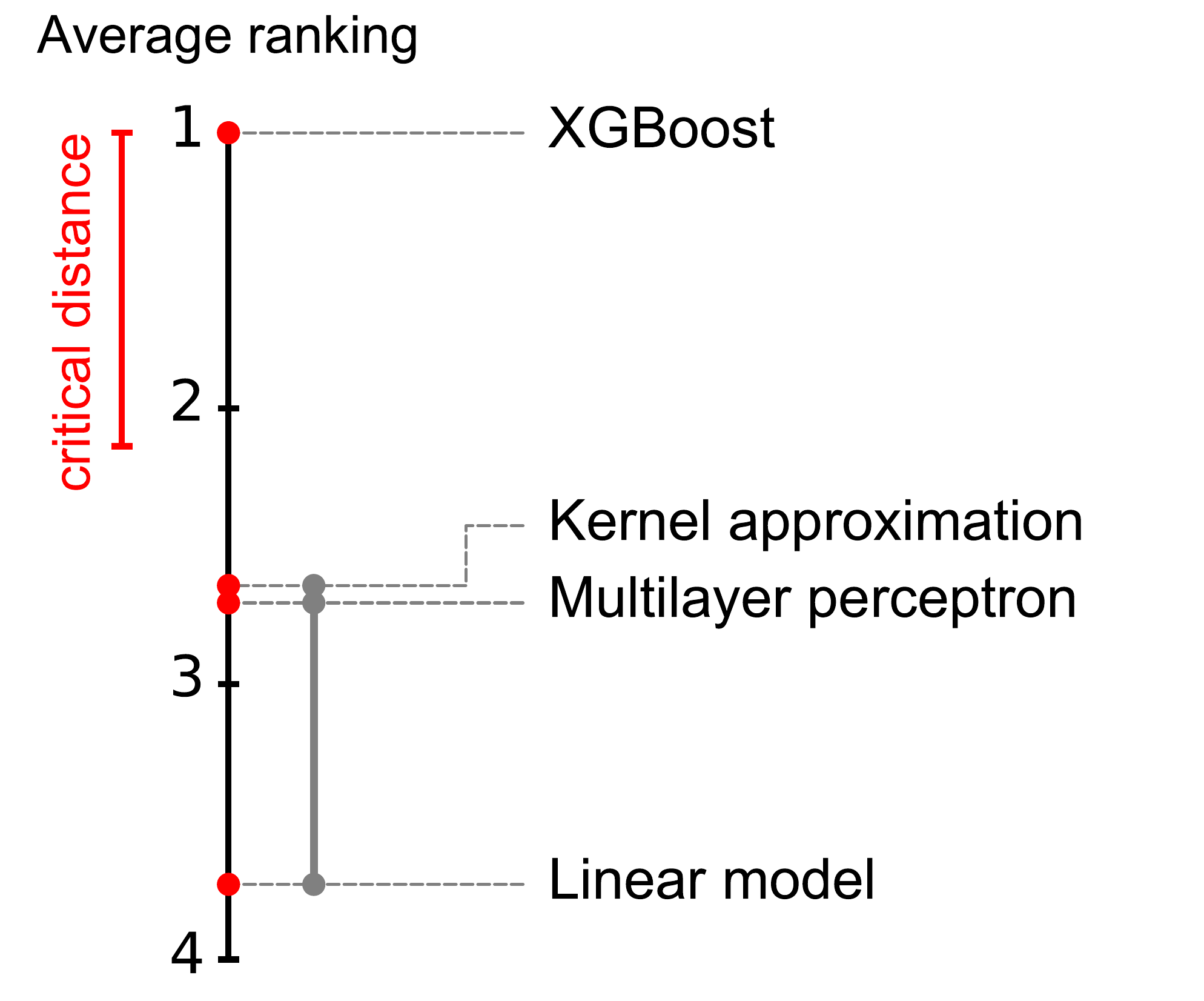}
		\caption{One-hot encoder}
	\end{subfigure}

	\vspace{.3cm}
	\begin{subfigure}[t]{.33\textwidth}
		\includegraphics[trim={-4cm 0cm 0 0cm},clip,height=.22\textheight]
		{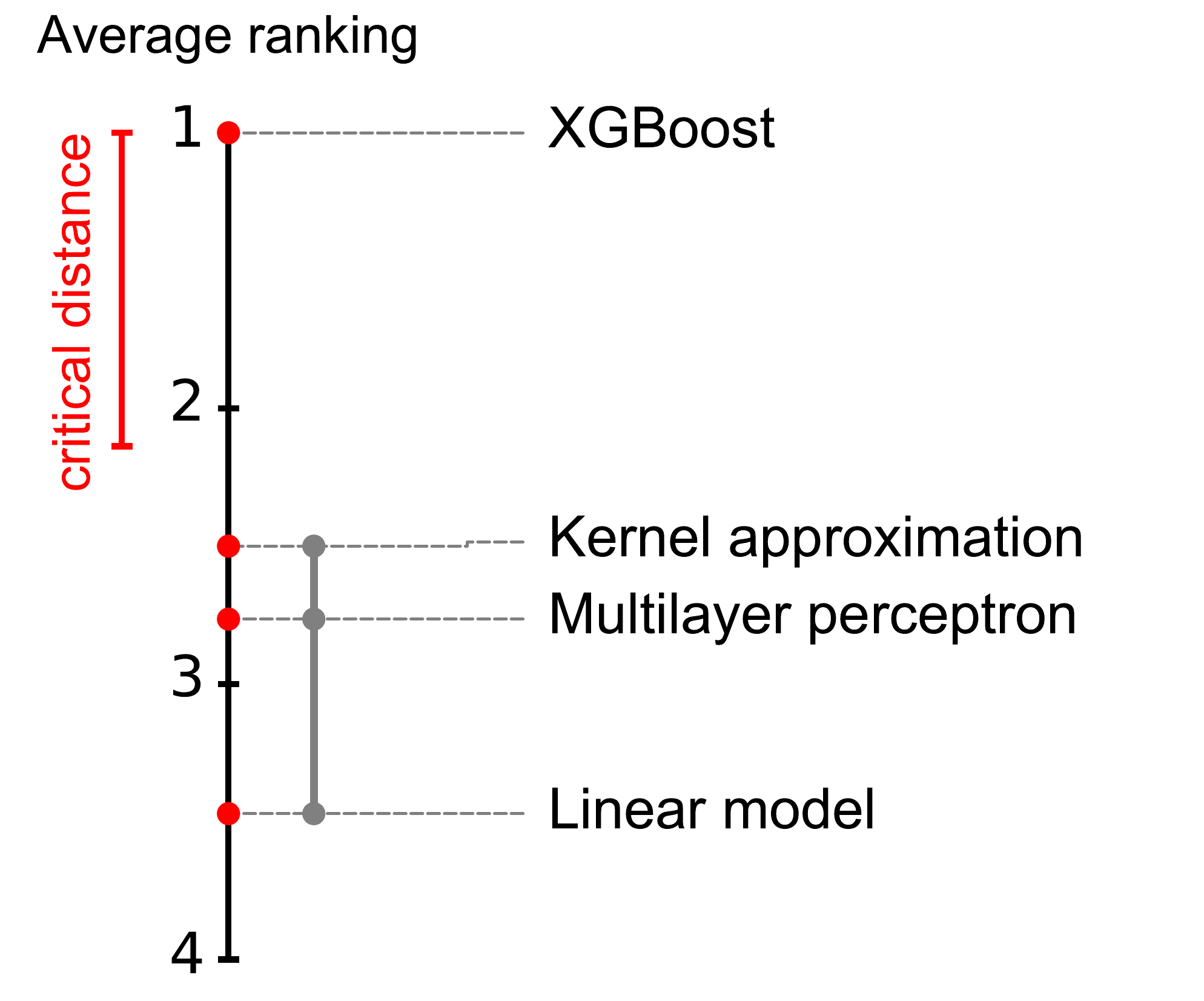}
		\caption{Gamma-Poisson factorization}
	\end{subfigure}

	\vspace{.3cm}
	\begin{subfigure}[t]{.33\textwidth}
		\includegraphics[trim={-4cm 0cm 0 0cm},clip,height=.22\textheight]
		{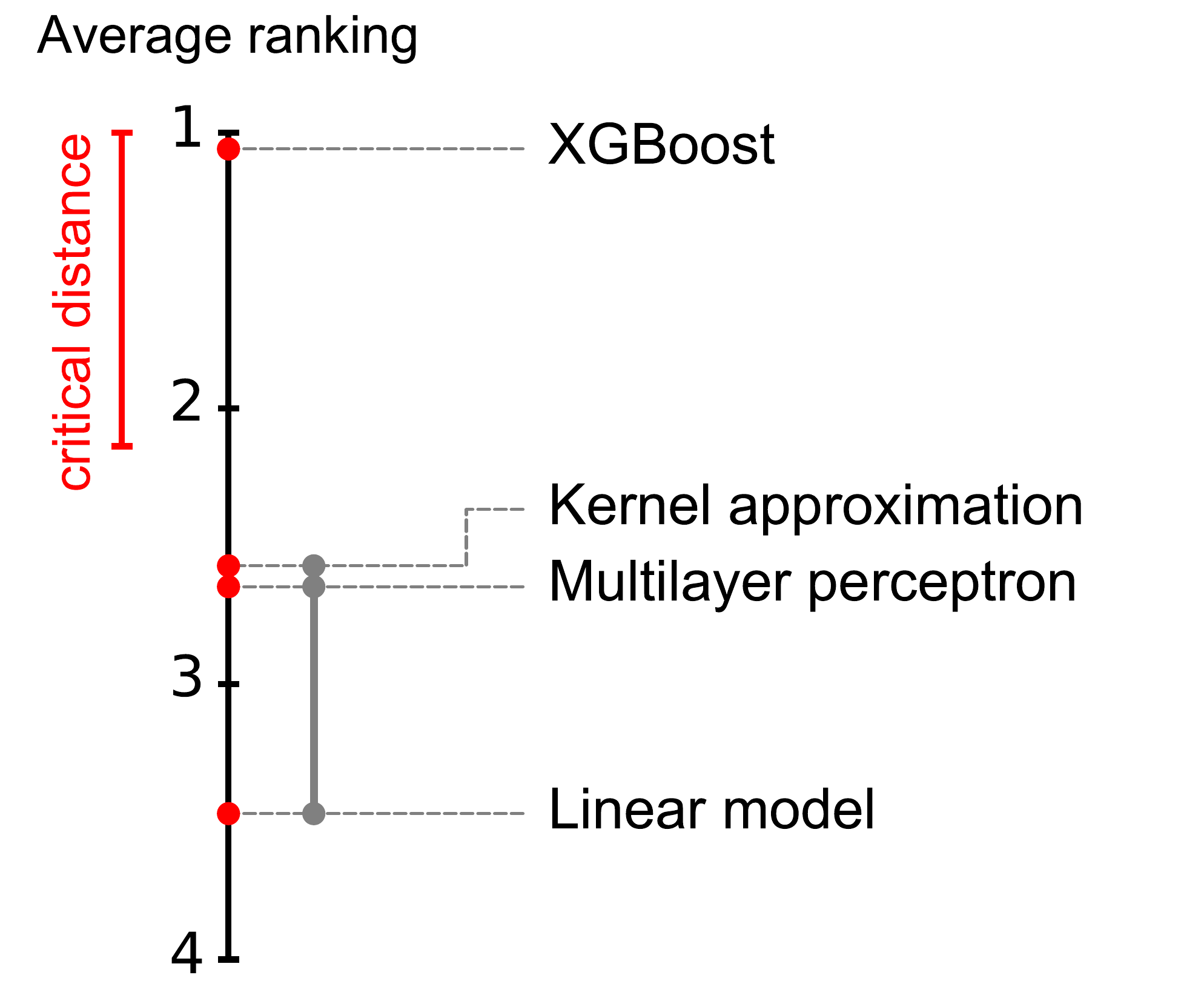}
		\caption{Min-hash encoder}
	\end{subfigure}%
	\caption{\textbf{Comparison of classifiers against each other with the
		Nemenyi post-hoc test}.
		Groups of classifiers that are not significantly different
		(at $\alpha$=0.05) are connected with a
		continuous gray line.
		The red line represents the value of the \emph{critical difference}
		for rejecting the null hypothesis. The benchmarked classifiers are:
		XGBoost; Polynomial kernel approx. with the Nystroem method, followed by an $\ell 2$ regularized linear/logistic regression (kernel approximation);
		a multilayer perceptron (1-2 layers); and a $\ell 2$ regularized linear/logistic regression (linear model).}
	\label{fig:nemenyi_classifiers} 
\end{figure}

	\vfill

	\begin{table}
		\caption{Median scores by dataset for XGBoost (d=30).}
		\label{tab:scores_datasets_XGB30}       
		\centering
		\scriptsize
		\setlength\tabcolsep{.1pt}
		\rowcolors{2}{black!5}{white}
		\begin{tabular}
			{L{.263\linewidth} C{.1\linewidth} C{.1\linewidth} C{.1\linewidth} C{.1\linewidth}
				C{.1\linewidth} C{.12\linewidth} C{.1\linewidth}}
			\hline\noalign{\smallskip}
			\textbf{Datasets} & Onehot SVD & Simila- rity enc. & TfIdf SVD & FastText SVD & Bert SVD &
			Gamma Poisson & Minhash encoder \\
			\hline\noalign{\smallskip}
			\input{scores_datasets_XGB_30.txt}
			\hline
		\end{tabular}
	\end{table}

	\begin{table}
		\caption{Median training and encoding times, in seconds, for Gamma-Poisson with
			XGBoost (d=30, a single fit, no hyper-parameter
			selection procedure).}
		\label{tab:times_GammaPoisson_XGB30}       
		\scriptsize
		\setlength\tabcolsep{5pt}
		\rowcolors{3}{black!5}{white}
		\begin{tabular}{lrrc}
			\hline\noalign{\smallskip}
			\multirow{2}{*}{\textbf{Datasets}} & Encoding time& Training time
			& Encoding time /\\
			 & Gamma-Poisson& XGBoost & training time\\
			\hline\noalign{\smallskip}
			\input{times_GaP_XGB_30.txt}
		    \hline
		\end{tabular}
	\end{table}

	\begin{table}[t!]
		\caption{\textbf{Recovering true categories for
				curated categorical variables.} NMI for different encoders ($d{=}|C|$).}
		\label{tab:nmi_encoders_clean_truedim}       
		\centering
		\footnotesize
		\setlength\tabcolsep{1.5pt}
		\rowcolors{3}{white}{black!5}
		\begin{tabular}{lccccc}
			\toprule
			\textbf{Dataset} & Gamma-   & Similarity & Tf-idf 	& FastText & Bert\\
			(cardinality) 	& Poisson & Encoding   &
			+ SVD 			& + SVD & + SVD \\
			\midrule
			{\bf Adult} (15) 				& \textbf{0.84} & 0.71 & 0.54 & 0.19 & 0.07	\\
			\textbf{Cacao Flavors} (100)             & \textbf{0.48} & 0.34 & 0.34 & 0.1 & 0.05		\\
			{\bf California Housing} (5) 	& \textbf{0.83} & 0.51 & 0.56  & 0.20 & 0.05	\\
			{\bf Dating Profiles} (19)  	& \textbf{0.47} & 0.26 & 0.29 & 0.12 & 0.06	\\
			{\bf House Prices} (15) 		& \textbf{0.91} & 0.25 & 0.32 & 0.11 & 0.05	\\
			\textbf{House Sales} (70) 		& \textbf{0.29} & 0.03 & 0.26 & 0.07 & 0.03	\\
			\textbf{Intrusion Detection} (66) & 0.27 & \textbf{0.65} & 0.61 & 0.13 & 0.06\\
			\hline
		\end{tabular}
	\end{table}

	\begin{table}[t!]
		\caption{\textbf{Recovering true categories for curated
				entries.} NMI for different encoders ($d$=10).}
		\label{tab:nmi_encoders_clean_d10}       
		\centering
		\footnotesize
		\setlength\tabcolsep{2.0pt}
		\rowcolors{3}{white}{black!5}
		\begin{tabular}{lccccc}
			\toprule
			\textbf{Dataset} & Gamma   & Similarity & Tf-idf 	& FastText &Bert\\
			(cardinality) 	& Poisson & Encoding   &
			+ SVD 			& + SVD & + SVD \\
			\midrule
			{\bf Adult} (15) 			 & \textbf{0.73} & 0.61 & 0.41 & 0.14 & 0.05 \\
			\textbf{Cacao Flavors} (100) & \textbf{0.44} & 0.28 & 0.21 & 0.05  &  0.03 \\
			{\bf California Housing} (5) 	& \textbf{0.63} & 0.51 & 0.56  & 0.20 & 0.05\\
			{\bf Dating Profiles} (19)  	& \textbf{0.34} & 0.28 & 0.20 & 0.08 & 0.03	\\
			{\bf House Prices} (15) 		& \textbf{0.81} & 0.26 & 0.26 & 0.09 & 0.04	\\
			\textbf{House Sales} (70) 		& \textbf{0.49} & 0.04 & 0.11 & 0.05 & 0.02	\\
			\textbf{Intrusion Detection} (66) & 0.34 & \textbf{0.53} & 0.46 & 0.08 & 0.04\\
			\hline
		\end{tabular}
	\end{table}
	
	\begin{table}[t!]
		\caption{\textbf{Recovering true categories for curated
				entries.} NMI for different encoders ($d$=100).}
		\label{tab:nmi_encoders_clean_d100}       
		\centering
		\footnotesize
		\setlength\tabcolsep{2.0pt}
		\rowcolors{3}{white}{black!5}
		\begin{tabular}{lccccc}
			\toprule
			\textbf{Dataset} & Gamma   & Similarity & Tf-idf 	& FastText &Bert\\
			(cardinality) 	& Poisson & Encoding   &
			+ SVD 			& + SVD & + SVD \\
			\midrule
			{\bf Adult} (15) 			 & 0.55 & \textbf{0.71} & 0.54 & 0.19 & 0.06 \\
			\textbf{Cacao Flavors} (100) & \textbf{0.47} & 0.34 & 0.34 & 0.10  &  0.05 \\
			{\bf California Housing} (5) & 0.18 & 0.51 & \textbf{0.56}  & 0.20 & 0.05 \\
			{\bf Dating Profiles} (19) 	 & \textbf{0.30} & 0.26 & 0.29 & 0.12 & 0.06 \\
			{\bf House Prices} (15) 	 & \textbf{0.63} & 0.25 & 0.32 & 0.11 & 0.05 \\
			\textbf{House Sales} (70) 	 & 0.21 & 0.03 & \textbf{0.26} & 0.07 & 0.03 \\
			\textbf{Intrusion Detection} (66) & 0.23 & \textbf{0.65} & 0.61 & 0.13 & 0.06 \\
			\hline
		\end{tabular}
	\end{table}

\end{document}

%% file: table_df_datasets.txt
\textbf{Crime Data          } &    1.5M &     135 &                \textbf{64.5} &        0.85 &            30.6 &                 Multi-label \\
\textbf{Medical Charges     } &    163k &     100 &                \textbf{99.9} &        0.23 &            41.1 &       	   Multi-label \\
\textbf{Kickstarter Projects} &    281k &     158 &               \textbf{123.8} &        0.64 &            11.0 &                 Multi-label \\
\textbf{Employee Salaries   } &    9.2k &     385 &               \textbf{186.3} &        0.79 &            24.9 &                 Multi-label \\
\textbf{Open Payments       } &    2.0M &    1.4k &               \textbf{231.9} &        0.90 &            24.7 &                 Multi-label \\
\textbf{Traffic Violations  } &    1.2M &   11.3k &               \textbf{243.5} &        0.97 &            62.1 &          Typos; Description \\
\textbf{Vancouver Employees } &    2.6k &     640 &               \textbf{341.8} &        0.67 &            21.5 &                 Multi-label \\
\textbf{Federal Election    } &    3.3M &  145.3k &               \textbf{361.7} &        0.76 &            13.0 &          Typos; Multi-label \\
\textbf{Midwest Survey      } &    2.8k &     844 &               \textbf{371.9} &        0.67 &            15.0 &               	 Typos \\
\textbf{Met Objects         } &    469k &   26.8k &               \textbf{386.1} &        0.88 &            12.2 &          Typos; Multi-label \\
\textbf{Drug Directory      } &    120k &   17.1k &               \textbf{641.9} &        0.81 &            31.3 &                 Multi-label \\
\textbf{Road Safety         } &    139k &   15.8k &               \textbf{790.1} &        0.65 &            29.0 &                 Multi-label \\
\textbf{Public Procurement  } &    352k &   28.9k &               \textbf{804.6} &        0.82 &            46.8 & Multi-label; Multi-language \\
\textbf{Journal Influence   } &    3.6k &    3.2k &               \textbf{956.9} &        0.10 &            30.0 & Multi-label; Multi-language \\
\textbf{Building Permits    } &    554k &  430.6k &               \textbf{940.0} &        0.48 &            94.0 &          Typos; Description \\
\textbf{Wine Reviews        } &    138k &   89.1k &               \textbf{997.7} &        0.23 &           245.0 &                 Description \\
\textbf{Colleges            } &    7.8k &    6.9k &               \textbf{998.0} &        0.02 &            32.1 &      	   Multi-label \\

%% file: scores_datasets_XGB_30.txt
building permits     &                 0.244 &              0.505 &             0.550 &             0.544 &             0.514 &            $\mathbf{0.570}$ &             0.566 \\
colleges             &                 0.499 &              0.532 &  $\mathbf{0.537}$ &             0.530 &             0.511 &                       0.524 &             0.527 \\
crime data           &                 0.443 &              0.445 &             0.445 &  $\mathbf{0.446}$ &             0.444 &                       0.445 &             0.446 \\
drug directory       &                 0.971 &              0.979 &             0.980 &  $\mathbf{0.982}$ &             0.979 &                       0.980 &             0.981 \\
employee salaries    &                 0.880 &              0.905 &             0.892 &             0.901 &  $\mathbf{0.913}$ &                       0.906 &             0.900 \\
federal election     &                 0.135 &              0.141 &             0.144 &  $\mathbf{0.151}$ &             0.147 &                       0.146 &             0.146 \\
journal influence    &                 0.019 &              0.138 &             0.164 &  $\mathbf{0.221}$ &             0.194 &                       0.118 &             0.133 \\
kickstarter projects &                 0.879 &              0.879 &             0.880 &  $\mathbf{0.880}$ &             0.880 &                       0.879 &             0.880 \\
medical charge       &                 0.904 &              0.905 &  $\mathbf{0.905}$ &             0.904 &             0.905 &                       0.904 &             0.904 \\
met objects          &                 0.771 &              0.790 &             0.789 &  $\mathbf{0.796}$ &             0.791 &                       0.791 &             0.794 \\
midwest survey       &                 0.575 &              0.635 &             0.646 &             0.636 &             0.605 &                       0.651 &  $\mathbf{0.653}$ \\
public procurement   &                 0.678 &              0.677 &             0.678 &             0.678 &             0.676 &                       0.674 &  $\mathbf{0.678}$ \\
road safety          &                 0.553 &              0.562 &             0.562 &             0.560 &             0.557 &                       0.563 &  $\mathbf{0.566}$ \\
traffic violations   &                 0.782 &              0.789 &             0.789 &             0.790 &             0.792 &                       0.792 &  $\mathbf{0.793}$ \\
vancouver employee   &                 0.395 &              0.550 &             0.530 &             0.509 &             0.506 &                       0.556 &  $\mathbf{0.568}$ \\
wine reviews         &                 0.439 &              0.671 &             0.724 &             0.657 &             0.669 &            $\mathbf{0.724}$ &             0.679 \\

%% file: times_GaP_XGB_30.txt
\bf building permits     &                      $1522$ &                         699 &                      $2.18$ \\
\bf colleges             &                          17 &                        $74$ &                      $0.24$ \\
\bf crime data           &                          28 &                      $1910$ &                      $0.01$ \\
\bf drug directory       &                         255 &                      $9683$ &                      $0.03$ \\
\bf employee salaries    &                           4 &                       $323$ &                      $0.01$ \\
\bf federal election     &                         126 &                       $764$ &                      $0.17$ \\
\bf journal influence    &                           7 &                        $18$ &                      $0.37$ \\
\bf kickstarter projects &                          20 &                       $264$ &                      $0.08$ \\
\bf medical charge       &                          42 &                       $587$ &                      $0.07$ \\
\bf met objects          &                         154 &                      $6245$ &                      $0.02$ \\
\bf midwest survey       &                           2 &                       $102$ &                      $0.02$ \\
\bf public procurement   &                         547 &                      $2150$ &                      $0.25$ \\
\bf road safety          &                         191 &                      $1661$ &                      $0.11$ \\
\bf traffic violations   &                         105 &                      $1969$ &                      $0.05$ \\
\bf vancouver employee   &                           2 &                         $9$ &                      $0.22$ \\
\bf wine reviews         &                      $1378$ &                         877 &                      $1.57$ \\